\documentclass[preprint,12pt]{elsarticle}

\usepackage{amssymb}

\usepackage{amsmath}
\usepackage{multirow}
\usepackage{subcaption}

\journal{Nuclear Physics B}

\begin{document}

\begin{frontmatter}



\title{An empirical study for the early detection of Mpox from skin lesion images using pretrained CNN models leveraging XAI technique}


\author[aff1]{Mohammad Asifur Rahim}
\author[aff1]{Muhammad Nazmul Arefin}
\author[aff2]{Md. Mizanur Rahman}
\author[aff2]{Md Ali Hossain}
\author[aff3,aff4]{Ahmed Moustafa}

\address[aff1]{Information and Computer Science, King Fahd University of Petroleum and Minerals(KFUPM), Dhahran, Saudi Arabia}
\address[aff2]{Faculty of Science and Information Technology, Daffodil International University, Dhaka, Bangladesh}

\address[aff3]{ Department of Human Anatomy and Physiology, The Faculty of Health Sciences, University of Johannesburg, Johannesburg, South Africa}

\address[aff4]{Centre for Data Analytics and School of Psychology, Bond University, Gold Coast, Queensland, Australia }

\begin{abstract}
\textbf{Context}: Mpox is a zoonotic disease caused by the Mpox virus, which shares similarities with other skin conditions, making accurate early diagnosis challenging. Artificial intelligence (AI), especially Deep Learning (DL), has a strong tool for medical image analysis; however, pre-trained models like CNNs and XAI techniques for mpox detection is underexplored. Objective: This study aims to evaluate the effectiveness of pre-trained CNN models (VGG16, VGG19, InceptionV3, MobileNetV2) for the early detection of monkeypox using binary and multi-class datasets. It also seeks to enhance model interpretability using Grad-CAM an XAI technique. \textbf{Method}: Two datasets, MSLD and MSLD v2.0, were used for training and validation. Transfer learning techniques were applied to fine-tune pre-trained CNN models by freezing initial layers and adding custom layers for adapting the final features for mpox detection task and avoid overfitting. Models performance were evaluated using metrics such as accuracy, precision, recall, F1-score and ROC. Grad-CAM was utilized for visualizing critical features.\textbf{ Results}: InceptionV3 demonstrated the best performance on the binary dataset with an accuracy of 95\%, while MobileNetV2 outperformed on the multi-class dataset with an accuracy of 93\%. Grad-CAM successfully highlighted key image regions. Despite high accuracy, some models showed overfitting tendencies, as evidenced by discrepancies between training and validation losses. \textbf{Conclusion}: This study underscores the potential of pre-trained CNN models in monkeypox detection and the value of XAI techniques. Future work should address dataset limitations, incorporate multimodal data, and explore additional interpretability techniques to improve diagnostic reliability and model transparency. 
\end{abstract}

\begin{graphicalabstract}

    \includegraphics[width=1\linewidth]{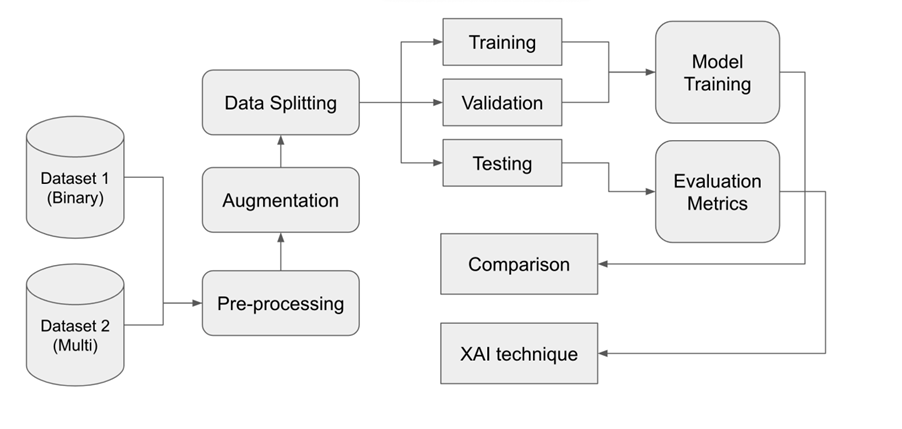}
    \captionof{figure}{Detailed methodology of our work}
    \label{fig:methodology}

     \includegraphics[width=1\linewidth]{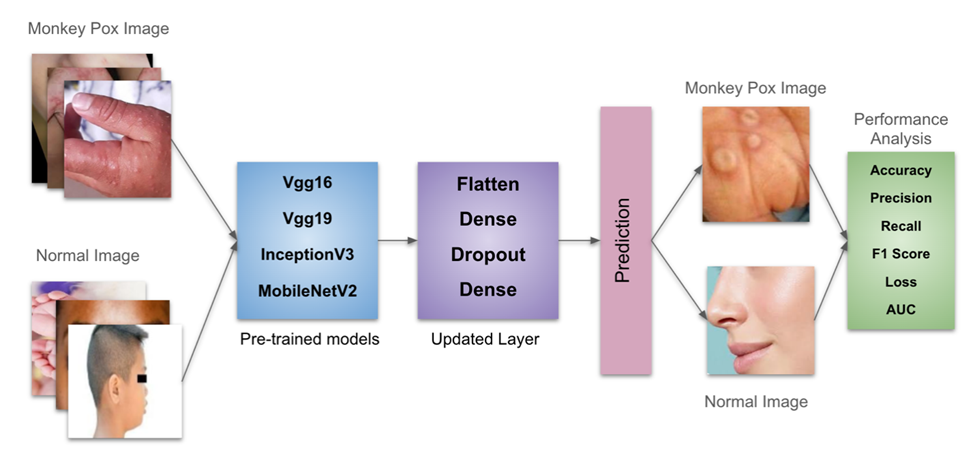}
    \captionof{figure}{Model Architecture}
    \label{fig:methodology}

\end{graphicalabstract}

\begin{highlights}
\item \textbf{Used robust yet lightweight pretrained CNN architectures:} 
    We used four well-established CNN models — VGG16, VGG19, InceptionV3, and MobileNetV2 — and fine-tuned them by freezing the initial layers and adding our custom layers. Now , this models were basically trained on huge dataset.Now, custom layers were added to adapt the final features to Mpox detection task and avoid overfitting. We also reduced the number of parameters to make the models more lightweight and efficient. This approach helped the models generalize effectively across two independent datasets.
    
    \item \textbf{Performed comprehensive comparative analysis:} 
    We performed detailed comparisons among all four models and compared our results with other state-of-the-art works. This transparent evaluation showed that our models achieved competitive performance, providing strong evidence for their clinical applicability.
    
    \item \textbf{Employed explainable AI (XAI) with Grad-CAM:} 
    We employed the Grad-CAM technique to visualize the important regions that influenced each prediction. This interpretability made the model’s decision-making more transparent and trustworthy for clinicians.

\end{highlights}


\end{frontmatter}



\section{Introduction}
\label{sec1}

\subsection{Motivation}
\label{subsec1}

Mpox is a viral disease that is transmitted from animals to humans (zoonotic), caused by the Mpox virus, which is part of the Orthodox virus genus \citep{bib1} of the Poxviridae family \citep{bib2}.  The genome of the Mpox virus (MPV), part of the Poxviridae family, is about 200 kb in length \citep{bib3}. It has conserved central regions responsible for encoding replication and assembly mechanisms \citep{bib3}. The terminal ends of the genome contain genes involved in pathogenesis and determining the host range \citep{bib3}. MPV has linear DNA \citep{bib3} and is typically recognized as a pleomorphic, enveloped virus with a dumbbell-shaped core and lateral bodies \citep{bib4}. Mpox is spread by direct contact with bodily fluids, skin lesions, or tiny respiratory droplets from infected animals, as well as by getting into contact with contaminated objects \citep{bib6}; \citep{bib5}. Mpox typically takes one to three weeks to incubate. During this time, non-specific clinical symptoms such as fever, enlarged lymph nodes, headaches, lethargy, and the development of skin lesions may appear.  \citep{bib7}. 
The first human case of monkeypox was documented in 1970 . It was observed in a newborn baby who developed a high fever followed by a rash on the face and body \citep{bib8}. The disease first appeared in the Congo and eventually spread throughout Africa, with a significant presence in Central and West Africa \citep{bib9}. Currently, this virus has spread outside of Africa. Since mid-2023, about 113 countries have reported 93,516 cases of the Monkeypox virus \citep{bib10}.  The World Health Organization (WHO) has voiced its concern over this disease, declaring it a global health emergency \citep{bib11}. Although the fatality rate is quite low (1-10) \% \citep{bib12}, It can cause severe medical conditions in some patients. Furthermore, no drug has yet been discovered that specifically targets the mpox virus.  That is why early detection of this disease is very important to stop it spreading. However, this disease is very difficult to diagnose because it has similarities with several skin-diseases such as: smallpox, roseola, etc. So, not early detection is essential but also accurately distinguishing between mpox and other similar skin diseases from skin lesion images is extremely important. Because it will help to take precautions and treat accordingly. 
In recent years, Artificial Intelligence (AI) has emerged as a powerful tool for analytical solutions \citep{bib13}. Machine learning (ML) which is a subfield of AI, particularly deep learning (DL), has proven to be highly effective in diagnosing diseases due to their low cost and energy efficiency \citep{bib14}.  Disease detection using deep learning (DL), a subfield of machine learning, has shown promise in image analysis and pattern recognition. Deep learning algorithms, when used in medical image analysis, can automatically detect and classify abnormalities in a range of medical images, such as X-rays, MRI scans, CT scans, and ultrasound images \citep{bib15}.  The Convolutional Neural Networks (CNN), a kind of deep learning model has been developed and found to be very effective in detecting several diseases from medical images \citep{bib16}.  In medical image analysis using deep learning algorithms, visual representations of convolutional neural networks (CNNs) illustrate a multi-layered structure. Early layers focus on detecting basic features like edges and textures, while deeper layers identify more complex and abstract patterns. This layered approach enables the network to automatically extract relevant information from medical images for tasks such as detection, segmentation, and classification \citep{bib17}. DL algorithms, including CNNs, typically require large amounts of data for training, leading to the issue of data scarcity \citep{bib18} .Additionally, CNNs tend to overlook long-range relationships within images, such as distant connections between objects \citep{bib19} .To address the data scarcity problem, transfer learning (TL) was introduced to achieve high performance on target tasks by leveraging knowledge learned from source tasks \citep{bib20} and the Vision Transformer (VIT) was developed, treating image classification as a sequence prediction task based on image patches, enabling it to capture these long-term dependencies . Recently, TL and VIT have also cross-pollinated the field of medical image analysis, where they are used for disease diagnosis and other clinical purposes \citep{bib17}\citep{bib18}. 
\subsection{Research Gap}
\label{subsec2}
There is a gap in proposing a model that combines a transfer learning strategy with learnable modules designed to make the models lightweight, interpretable, and generalizable across multiple datasets.

\subsection{Contribution}
\label{subsec3}
\begin{itemize}
    \item \textbf{Used robust yet lightweight pretrained CNN architectures:} 
    We used four well-established CNN models — VGG16, VGG19, InceptionV3, and MobileNetV2 — and fine-tuned them by freezing the initial layers and adding our custom layers. Now , this models were basically trained on huge dataset.Now, custom layers were added to adapt the final features to Mpox detection task and avoid overfitting. We also reduced the number of parameters to make the models more lightweight and efficient. This approach helped the models generalize effectively across two independent datasets.
    
    \item \textbf{Performed comprehensive comparative analysis:} 
    We performed detailed comparisons among all four models and compared our results with other state-of-the-art works. This transparent evaluation showed that our models achieved competitive performance, providing strong evidence for their clinical applicability.
    
    \item \textbf{Employed explainable AI (XAI) with Grad-CAM:} 
    We employed the Grad-CAM technique to visualize the important regions that influenced each prediction. This interpretability made the model’s decision-making more transparent and trustworthy for clinicians.
\end{itemize}

The organization of the paper is structured as follows: Section 2 reviews the related works, highlighting prior research and identifying gaps in monkeypox detection using deep learning techniques. Section 3 outlines the design of the experiment, detailing objectives, research questions, and methodology. Section 4 describes the datasets used, including their composition, preprocessing, and augmentation processes. Section 5 covers the execution of the experiment, explaining model selection, hyperparameter tuning, and training processes, while Section 6 presents the results and analysis, comparing model performance across binary and multi-class datasets and discussing interpretability using Grad-CAM. Section 7 offers a discussion of findings, contextualizing them within existing literature and practical applications. Section 8 addresses threats to validity, highlighting potential limitations in the methodology and outcomes. Finally, Section 9 concludes the paper with future work suggestions and a summary of key contributions.

\section{Related Work}
\label{sec2}
Research over the past few years has been done on the detection of Monkey Pox from skin lesion images. ML and DL models were found to be extremely effective in this regard.   
A model called “PoxNet22” a modified inceptionV3 was proposed in \citep{bib21}. Transfer learning techniques were applied for feature extraction. Moreover, to handle overfitting data augmentation was utilized. The proposed model achieved a 99\% recall and precision. However, they did not validate their model in multiple datasets which raises questions about the generalizability of their proposed model. Moreover, they did not use XAI techniques to interpret their model's performance \citep{bib21}. 
\cite{bib22} utilized several CNN-based pre-trained models and vision transformer(vit) for the classification purpose of monkeypox in binary classification datasets.  They achieved an accuracy of 93 and 99 accuracy respectively in two datasets. On top of that, for the transparency of the models, Local Interpretable Model Agnostic Explanations (LIME) was utilized. However, they only used binary classification dataset not multi class data 
Attention-based MobileNetV2 was proposed by \citep{bib23}. To get rapid and better performance, both spatial and channel attention mechanisms are tailored. They achieved an accuracy of 98.19\% on MSID dataset which is publicly available. Gradient-weighted Class Activation Mapping (Grad-CAM) and Local Interpretable Model-Agnostic Explanations (LIME) were used for model’s interpretability and transparency.  
One of the research projects conducted \citep{bib24}, applied 13 pre-trained DL models using transfer learning on a Monkeypox dataset and identified the best-performing models to create an ensemble for improved detection performance. The ensemble achieved the highest accuracy (87.13\%) and F1-score (85.40\%), while Xception was the second-best individual model. However, the dataset size was small, and reliance on pre-trained models limits deployment in memory-constrained settings, necessitating lightweight model designs. Moreover, XAI techniques are also not applied.
\cite{bib25} proposed a modified DenseNet201 model using a dataset of three images of skin diseases. Additionally, Gradient-weighted Class Activation Mapping (Grad-Cam) and Local Interpretable Model-Agnostic Explanations (LIME) were created to make the results more transparent and understandable. However, they did not include execution time of their model, so it is unclear whether their proposed model is memory efficient. For real-time detection of monkeypox images a modified YOLOv5 model was proposed by \citep{bib26}. Using three different hyperparameters tuning strategies, they achieved an accuracy of 98.18\%. They used the transfer learning strategy. However, the dataset size was small . 
According to \citep{bib27}, a Vision Transformer and CNN-based ensemble achieved an accuracy of 81.91\% on a dataset comprising seven skin lesion categories, including monkeypox. However, the study was conducted on a small dataset, which limited the model's performance. Additionally, no XAI techniques were employed in their research.To solve the problem of data-imbalance and complexities in skin images, two parallel CNN architectures along with attention mechanisms were proposed \citep{bib28}. They developed an effective attentive mechanism process through transfer learning that leads to providing better discriminative feature maps. However, the main limitation of this study is that they did not interpret the performance of their CNN models using XAI techniques. 
This study was conducted to overcome key challenges in monkeypox detection, such as the extraction of irrelevant features from low-contrast images, high memory and computational complexity, and the need for large datasets to prevent overfitting. To address these limitations, the MOX-NET algorithm is introduced \citep{bib29}. The approach begins with a fusion-based contrast enhancement algorithm to refine image quality. Deep features are then extracted using six modified DL architectures—ViT, Swin Transformers, ResNet-50, ResNet-101, EfficientNetV2, and ConvNeXt-V2—trained via transfer learning. These features are integrated using a CSID fusion strategy, and an ECF-based method meticulously selects optimal features for recognition. Finally, an M-SVM classifier is employed to classify the features, achieving superior classification accuracy and efficiency. However, no interpretable techniques were applied to enhance the reliability of the model.
In \citep{bib30} a modified DenseNet-201 model is proposed and used it in a dataset of original images and augmented images consisting of six classes, including monkeypox virus. They achieved an accuracy of 93.19 \% and 98.91\% accuracy respectively. Moreover, they also compared their results with other states of the art. Finally, they used Grad-Cam to interpret the performance of their model.  However, they only used one dataset to validate the model which raises the threat of external validity.
In \cite{bib31}, ensemble approach is proposed which consists of three deep learning pre-trained CNN architectures . On top of that, particle swarm optimization (PSO) algorithm is utilized to assign optimized weights to each model during the ensemble process. Their model achieved an accuracy of 97.78\% in a publicly available four-class dataset. Moreover, Grad-Cam techniques were also utilized. However, the limitations remain that they did not use additional dataset for validation purposes which raises external threat.
\cite{bib32} focuses on distinguishing monkeypox lesions from other types of skin lesions for rapid and accurate detection. To achieve this, a combined dataset was created by merging two datasets, comprising seven types of skin lesions, including monkeypox. They utilized several pretrained CNN models and achieved an accuracy of 74.76\% which is quite low. Moreover, Interpretable techniques were not applied.\\
By analyzing the papers, several research gaps have been found: The first one is that very few researchers have proposed a model and validated it with binary and muti datasets. Moreover, XAI techniques for interpreting the model’s performance have not been explored that much. Additionally, how transfer learning techniques reduce complexity are not analyzed properly. So, in this paper, we aim to use pre-trained CNN models for the early detection of monkeypox from skin lesion images. Not only that, but a model will also be proposed and explained how transfer learning reduces the complexity. Moreover, overfitting issues will be discussed thoroughly using accuracy-loss curves. Finally, a modern XAI technique called Gradient-weighted Class Activation Mapping (GRAD-CAM) will be applied to visualize and understand the decisions made by CNN.

\section{Methodology}

\subsection{Design of the experiment}

\textbf{Purpose}: To analyze and evaluate the effectiveness of pre-trained CNN models for early detection of monkeypox from skin lesion images.
\textbf{Issue}: By accurately classifying monkeypox cases in both binary and multiclass datasets and identifying key visual features relevant to early detection.
\textbf{Object}: Using model performance metrics to assess detection accuracy and other metrics and applying explainable AI (XAI) techniques to highlight critical image features for identification.
\textbf{Viewpoint}: From the viewpoint of researchers and healthcare professionals interested in accurate early detection and feature relevance for diagnosis.
So, according to Goal Question metric (GQM) \citep{bib33}, The main objective is to analyze and evaluate the effectiveness of pre-trained CNN models for early detection of monkeypox from skin lesion images, using model performance metrics to assess detection accuracy and XAI techniques to highlight important features for identification, from the viewpoint of researchers and healthcare professionals aiming to improve early diagnostic accuracy and understanding of key visual indicators.

\subsubsection{Treatments, Dependent and Independent variables}

\textbf{Treatments}: It refers to the different interventions or conditions being tested to observe their impact. In our work, since we are evaluating the effectiveness of pre-trained CNN models such as: VGG16, VGG19, InceptionV3 and MobileNetV2, these models serve as the treatments.

\textbf{Dependent variables}: Dependent variables are the outcomes or measurements used to assess the effects of the independent variables. In this study, they include model performance metrics such as accuracy, precision, recall, F1-score, AUC, and explainability outputs, specifically the heatmaps generated by XAI techniques like Grad-CAM.

\subsubsection{Research Questions}
In this empirical study the following research questions are answered:\\
\textbf{RQ1}: Which pre-trained CNN models (VGG16, VGG19, InceptionV3, MobileNetV2) demonstrate the best performance in detecting monkeypox from skin lesion images based on key evaluation metrics (accuracy, precision, recall, F1-score, execution time)?\\
To answer this RQ, each pretrained model trained and evaluated using different evaluation metrics in two datasets of skin lesion images of to detect monkeypox virus and the best performing model will be identified by comparative analysis among those models. This RQ is addressed on 4. Experiment Results.\\
\textbf{RQ2}: Which technique can be applied to reduce the complexity of the CNN model? \\
In this RQ, the proposed model's architecture settings are discussed. This RQ is addressed in 3.3 Execution of Experiment.\\
\textbf{RQ3}: Which hyperparameter settings can be applied to train the model efficiently?\\
\textbf{RQ4}: Which technique can be applied to improve the interpretability of the model’s performance?\\
To answer this RQ, explainable AI(XAI): GRAD-CAM techniques are adopted. The outputs are generated by highlighting features inside the images by heatmap. This section is addressed in Experimental Result.\\

\begin{figure}
    \centering
    \includegraphics[width=1\linewidth]{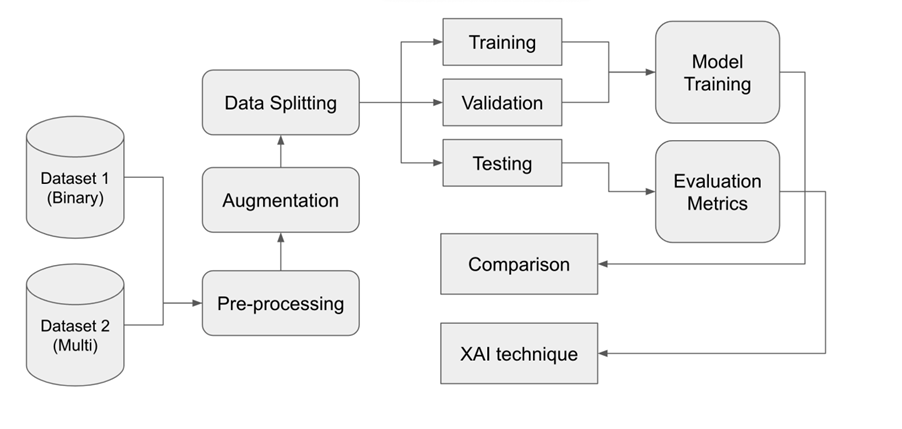}
    \caption{Detailed methodology of our work}
    \label{fig:methodology}
\end{figure}

In Figure \ref{fig:methodology}, the detailed methodology of our work is illustrated. Two datasets were utilized to validate the performance of the proposed model. The first dataset contained two classes: Monkeypox and Others, while the second dataset comprised six classes. Both datasets were preprocessed through resizing and normalization. Subsequently, data augmentation techniques were applied to increase the number of images and enhance diversity. The datasets were then split into training, validation, and test sets. Four pretrained models—VGG16, VGG19, InceptionV3, and MobileNetV2—were fine-tuned by integrating our proposed custom layers. The performance of these models was evaluated on the test data using various evaluation metrics. Finally, Grad-CAM was employed to visualize and highlight the key features of the images that contributed to the model’s predictions. The detailed methodology of our work is explained in Proposed model architecture. \\
In this study, only augmented images were used, as the main dataset has not enough image data for the fine tuning the models. Limited data is a common issue in specialized field like medical imaging \citep{bib34}. Augmentation forces the model to recognize the pattern despite the transformations. Taylor et al. showed in \citep{bib35} that models trained on augmented data are more resilient to input distortions and transformations.

\subsection{Dataset Description}

In this experiment, two distinct datasets were utilized to develop a detection system for monkeypox using skin lesion images. Both datasets were from Kaggle. Those datasets provided essential visual data points to train machine learning models efficiently. \\
\textbf{Monkeypox Skin Lesion Dataset}
The first dataset, titled Monkeypox Skin Lesion Dataset by Nafisa \citep{bib50} and includes approximately 400 images. However, there is another augmented version of this dataset which includes 3192 images. In this paper only the augmented version is used. This dataset has two classes. The augmented dataset distribution is given Table \ref{table1}. This dataset contains a diverse set of skin lesion images that capture various stages of monkeypox infections. The images have different resolutions, which creates complexity to preprocessing. However, this provides a more varied input for training machine learning models. Basic labels are provided in this dataset, distinguishing between monkeypox-affected and non-affected skin.\\ 
\textbf{Mpox Skin lesion Dataset Version 2.0}
The second dataset, known as Mpox Skin lesion Dataset Version 2.0 (MSLD v2.0) was developed by Nafisa \citep{bib49} and consists of over 1000 images. It is significantly larger than the first dataset and contains good quality images that simplify the preprocessing. There is also another augmented version of this dataset. The augmented version of this dataset is used in this study. The augmented version are divided into five folds.We used the fifth fold only that contains about 7532 images to save computational overhead .  A key feature of MSLD v2 is its comprehensive set of detailed annotations. This includes additional metadata such as lesion type and demographic indicators. This labeling facilitates the creation of more classification models that can incorporate additional data features to enhance accuracy. In Table \ref{table2} detail distribution of the MSLD v2.0 is given. 



\begin{table}[h!]
\centering
\caption{Detail distribution of the MSLD}
\begin{tabular}{l c} 

 \hline
 Class Label & No. of Images \\ 
 \hline
 Monkey-Pox & 1428 \\ 
 Others     & 1764 \\ 
 \hline
 Total      & 3192 \\ 
 \hline
\end{tabular}

\label{table1}
\end{table}

\begin{table}[h!]
\centering
\caption{Detail distribution of the MSLD v2.0}
\begin{tabular}{l c}
\hline
Class Label & No. of Images \\
\hline
Mpox & 2968 \\
Chickenpox & 742 \\
Measles & 532 \\
Cowpox & 602 \\
Hand, foot, and mouth disease & 1526 \\
Healthy & 1162 \\
\hline
Total & 7532 \\
\hline
\end{tabular}

\label{table2}
\end{table}

\begin{figure}
    \centering
    \includegraphics[width=1\linewidth]{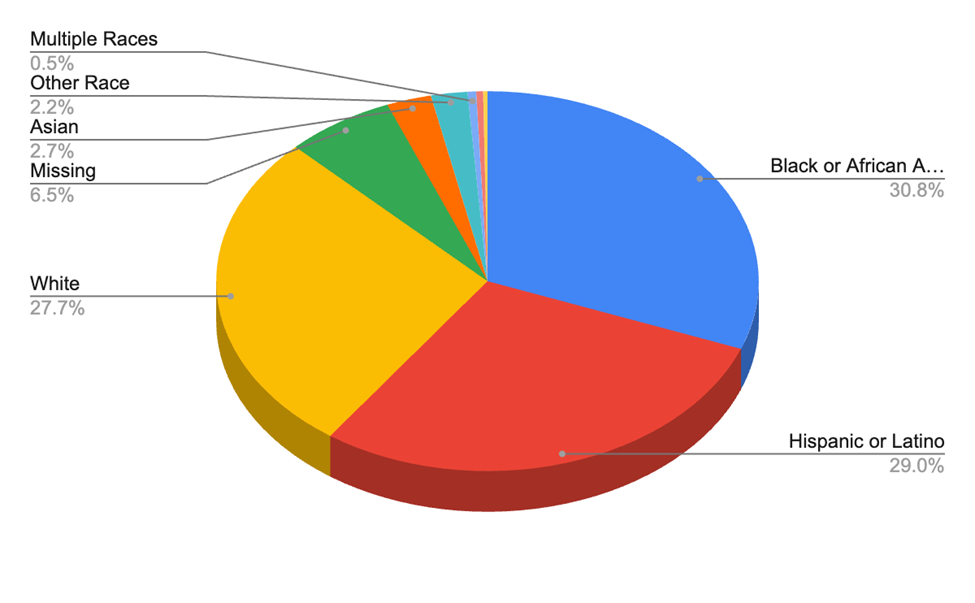}
    \caption{Race wise Mpox distribution}
    \label{fig:Race_Wise_mpox}
\end{figure}

\begin{table}[h!]
\centering
\caption{Comparison between Monkeypox Skin Lesion Dataset and Mpox Skin Lesion Dataset Version 2.0.}
\begin{tabular}{|l|p{1.5cm}|p{1.5cm}|}
\hline
\textbf{Attribute} & \textbf{Mpox Skin Lesion Dataset} & \textbf{Mpox Skin Lesion Dataset Version 2.0} \\

\hline
Class & 2 & 6 \\
\hline
Number of Original Images & $\sim$400 & $>$1000 \\
\hline
Number of Augmented Images & $\sim$1200 & $>$3000 \\
\hline
Image Resolution & Varying & Consistent \\
\hline
Annotation & Basic labels & Detailed, includes metadata \\
\hline
Diversity & Limited & High \\
\hline
\end{tabular}

\label{table3}
\end{table}

According to \citep{bib36}, the Mpox cases by race and ethnicity indicates that, most cases are among Black or African American individuals. 30.80\% of reported cases include people in this race. This is closely followed by Hispanic, who represent 29.04\%, white individuals account for 27.65\% of the cases. Asian individuals at 2.66\%, and rest at 2.18\%. Figure \ref{fig:Race_Wise_mpox} given the overall view of the percentage of the distribution of Mpox cases by race and ethnicity. 

Table \ref{table3} compares two datasets utilized for monkeypox detection. The first dataset, Monkeypox Skin Lesion Dataset, contains two classes and approximately 400 original images, which were augmented to around 1,200 images. The dataset has varying image resolutions and provides only basic labels for classification. Its diversity is limited, which may restrict the model’s ability to generalize across different scenarios. In contrast, the Mpox Skin Lesion Dataset Version 2.0 is more comprehensive, with six classes and over 1,000 original images that were augmented to exceed 3,000 images. This dataset offers consistent image resolution, detailed annotations, and metadata that facilitate advanced analysis and model training. Its high diversity makes it more suitable for developing models with better generalization capabilities.

\begin{figure}
    \centering
    \includegraphics[width=01\linewidth]{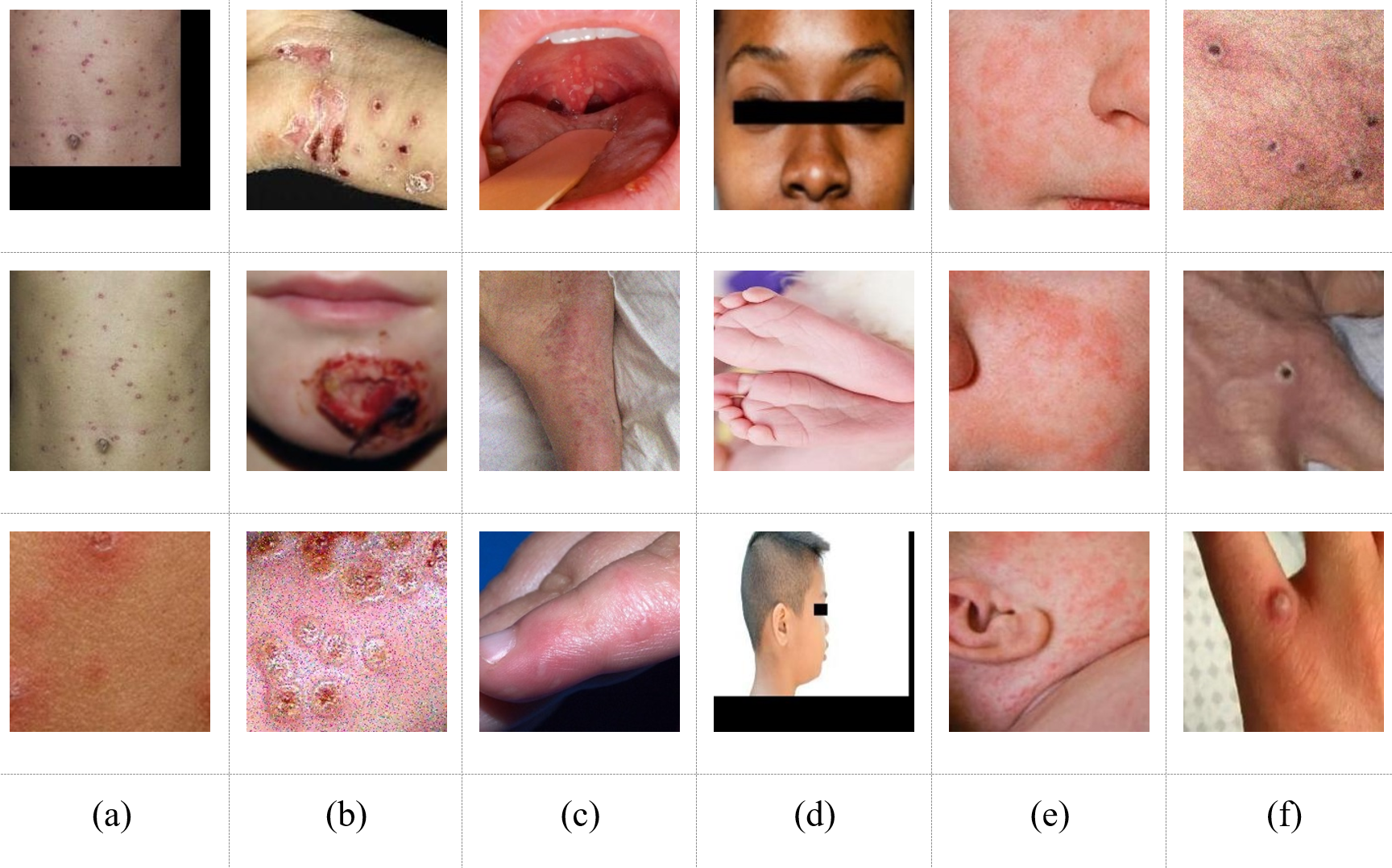}
    \caption{Mpox Skin Lesion Dataset V2 (MSLD v2.0) (a) Chickenpox, (b) Cowpox, (c) HFMD, (d) Healthy, (e) Measles, (f) Mpox}
    \label{fig:enter-label1}
\end{figure}

\begin{figure}
    \centering
    \includegraphics[width=01\linewidth]{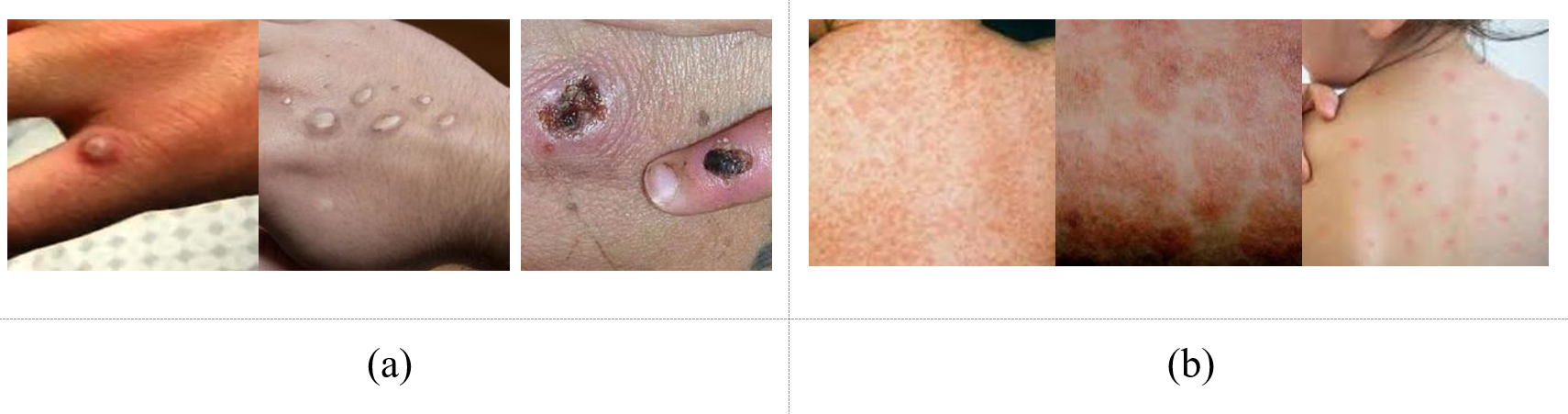}
    \caption{Monkeypox Skin Lesion Dataset (MSLD) (a) Monkeypox, (b) Others}
    \label{fig:enter-label2}
\end{figure}
Figure \ref{fig:enter-label1} and \ref{fig:enter-label2} displays some sample images of each dataset

\subsection{Execution of Experiment}
\textbf{RQ2}: Which technique can be applied to reduce the complexity of the CNN model?\\
\textbf{RQ3}: Which hyperparameter settings can be applied to train the model efficiently?
In this section, the RQ2 and RQ3 is answered in detail.
\subsubsection{Experimental Setup}
The experimental setup for this study was implemented on Google Colab, utilizing Python 3 as the runtime environment and a T4 GPU as the hardware accelerator, which enabled efficient training and fine-tuning of pre-trained CNN models on large image datasets. TensorFlow and Keras were employed as the primary libraries for building, training, and evaluating the models due to their flexibility and ease of use in deep learning tasks. Google Drive was used for storing datasets and saving model checkpoints, while visualization tools like Matplotlib were used for plotting training curves, confusion matrices, and evaluation metrics. This setup allowed for effective experimentation and resource management throughout the study.
\subsubsection{Data preparation}
\textbf{Data preprocessing}: After loading dataset to the Google colab, first thing to check whether images are correctly labeled and organized or not. According to the quality of the images, resize operation was done to fixed dimensions (e.g., 224x224x3) for model compatibility. Similarly, normalizing pixel values to standardize input for neural networks was also employed.

\textbf{Data Splitting}: In both datasets, the data is divided into training, validation, and test sets in the ratios of 75\%, 15\%, and 10\%, respectively. For the binary dataset, 3,192 images are split into 2,394 images for training, 478 images for validation, and 320 images for testing. Similarly, for the multi-class dataset, 7,532 images are split into 5,649 images for training, 1,129 images for validation, and 754 images for testing. Table \ref{table4} and Table \ref{table5} gives the data distribution of both datasets after splitting .

\begin{table}[h!]
\centering
\caption{Data distribution after splitting for MSLD.}
\begin{tabular}{|l|c|c|c|c|}
\hline
\textbf{Class-name} & \textbf{Training} & \textbf{Validation} & \textbf{Testing} & \textbf{Total} \\
\hline
Monkey-pox & 1071 & 214 & 143 & 1428 \\
Others     & 1323 & 264 & 177 & 1764 \\
\hline
Total      & 2394 & 478 & 320 & 3192 \\
\hline
\end{tabular}

\label{table4}
\end{table}

\begin{table}[h!]
\centering
\caption{Data distribution after split for MSLDv.2.}
\begin{tabular}{|l|c|c|c|c|}
\hline
\textbf{Class-name} & \textbf{Training} & \textbf{Validation} & \textbf{Testing} & \textbf{Total} \\
\hline
Monkey-Pox & 2226 & 445 & 297 & 2968 \\
HFMD       & 1145 & 228 & 153 & 1526 \\
Healthy    & 871  & 174 & 117 & 1162 \\
Chickenpox & 557  & 111 & 74  & 742  \\
Cowpox     & 451  & 91  & 60  & 602  \\
Measles    & 399  & 80  & 53  & 532  \\
\hline
Total      & 5649 & 1129 & 754 & 7532 \\
\hline
\end{tabular}

\label{table5}
\end{table}

\subsubsection{Data Augmentation} 
To enhance the training process and mitigate potential issues related to overfitting, image augmentation techniques were applied to the datasets. Basic augmentation techniques like rotation, flipping, zooming, and brightness adjustments to artificially expand the dataset’s size and variability. The use of augmented images contributed to improving the model’s performance, especially when dealing with limited original data. Table 3 compares the key attributes of the two datasets, including augmentation.
\subsubsection{Model Selection and Training}
\textbf{VGG-16 (Visual Geometry Group 16)}
VGG16 \citep{bib37} is a CNN architecture known for its simple structure (3X3) convolutional filters throughout the entire network. It contains 16 layers with learnable parameters, where the name “16” comes from. This layer contains 13 convolutional layers and 3 fully connected layers for classifications. It uses 3x3 filters in all convolutional layers and applies max polling to reduce spatial dimensions. VGG16 is effective at feature extraction while it achieves high accuracy. Its depth and large number of parameters made it computationally demanding.\\  
\textbf{VGG19 (Visual Geometry Group 19)}
VGG-19 \citep{bib37} contains 19 layers with learnable parameters, e.g., 16 conv layers and 3 fully connected layers. It is using 3x3 conv filters with stride of 1 and padding of 1 for spatial resolution maintenance. It also includes max-pooling layers of 2x2 with stride 2 and ends with 3 fully connected layers and a softmax for classification. Every conv layer is followed by a ReLU activation function to introduce non-linearity. The small filter size reduces the number of parameters compared to larger filters, making it more computationally efficient. The network accepts images of size 224x224x3 as input. Images are resized and normalized before feeding into the network. \\
\textbf{InceptionV3}
Inception V3 \citep{bib39} builds upon earlier version of the Inception architecture, particularly inception V1 (GoogLeNet) and V2. It combines Inception modules with various filter size (1x1, 3x3, 5x5, etc.) to take multi scale features in the layer. It also includes auxiliary classifiers during training to improve convergence. In the core improvement, it replaces larger convolutions, e.g., 5xt with multiple smaller conv, e.g., two 3x3 to reduce computational complexity. Additionally, it splits 3x3 conv into two 1xn and nx1 conv to save computation and introduced careful pooling strategies to down-sample feature maps without unnecessary information loss. It processes images of size 299x299x3 (larger than standard size like 224x224 for VGG) and achieved high accuracy on the ILSVRC dataset. \\
\textbf{MobileNetV2}
MobileNet V2 \citep{bib38} is a lightweight CNN architecture designed for mobile and embedded vision applications. It is an improved version of MobileNet V1 that emphasizes performance for devices with limited resources. Like MobileNetV1 it uses depth wise separable convolutions which splits convolution into two operations, i.e., a single filter per input channel and 1x1 conv to combine outputs from depth-wise conv. It contains an initial convolution layer followed by a series of bottleneck layers ended with fully connected layer for classification. Typically accepts images of size 224x224x3 but can handle smaller resolution for faster interface. 

\subsubsection{Proposed model architecture}
\begin{figure*}
    \centering
    \includegraphics[width=0.75\linewidth]{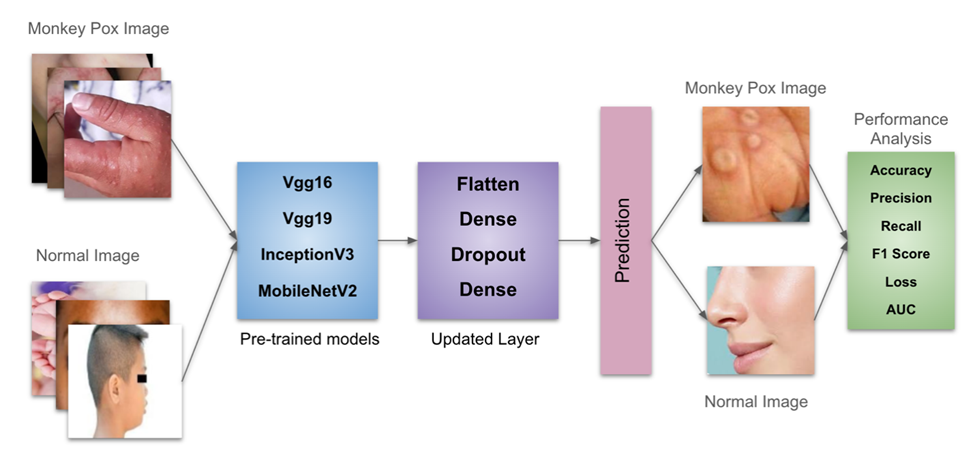}
    \caption{Model architecture}
    \label{fig: Model_architecture}
\end{figure*}
\begin{figure}
    \centering
    \includegraphics[width=1\linewidth]{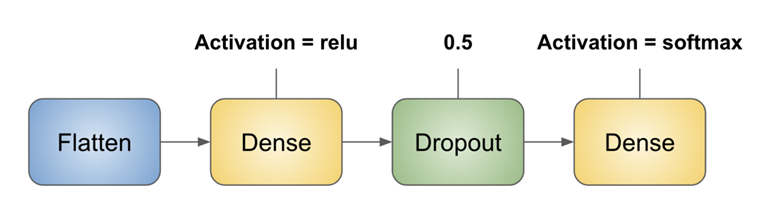}
    \caption{Modified Layer}
    \label{fig: modified_layer}
\end{figure}

Figure \ref{fig: Model_architecture} displays the overall proposed model architecture and Figure \ref{fig: modified_layer} displays the added customized layers.
A pre-trained model refers to the model that has been training for datasets on several fields and can be reused for similar datasets. Transfer learning techniques has been adopted in our study to train the models. The model has been trained using the large Monkey Pox dataset, which comprises hundreds of photos in many categories. This enables it to identify characteristics in various pictures, including images of monkey pox in various body parts like hands and face. Moreover, regardless of the relatively small dataset, transfer learning can allow the use of the learned information from the existing model to improve the performance of specific tasks. Numerous works have demonstrated the benefits of pre-trained models and their improved performance \citep{bib40}. The model architecture contains following six steps:
\begin{enumerate}[1.]
    \item Initially, the base models have been initialized with a pre-trained network that lacks fully connected layers.
    \item To preserve the knowledge learned during ImageNet training, all layers of the base model were frozen. This ensured that the feature extraction capability of the models remained intact and was not modified during training.
    \item A Flatten Layer was added to transform the feature map output of the base model into a 1D array suitable for dense layer input.
    \item A Dense Layer with 256 neurons and ReLU activation was incorporated to learn complex patterns specific to the new dataset.
    \item A Dropout Layer with a rate of 0.5 was included to reduce overfitting by randomly deactivating neurons during training.
    \item Finally, a Dense Output Layer with neurons (equal to the number of classes) and softmax activation was added for classification.
\end{enumerate}

In this study, all models follow the modification shown in Figure \ref{fig: modified_layer}. Pre-trained methods were previously used for image classification; however, our work focuses only on mpox detection.In this study four well-known CNN architectures – VGG16 , VGG19, InceptionV3, MobileNetV2 are used to evaluate the performance on MSLD v2 and MSLD datasets. From the previous works, it is seen that those models performed excellent interims of classify images through transfer learning in several image analysis. Using strong models like VGG16 and VGG19 to lightweight models e.g., MobileNetV2 the study evaluated CNN architectures to identify weakness and strength of each model for mpox detection. Transfer learning includes the training deep learning model on a large dataset and then utilizes the trained model’s parameters to adjust training on a smaller data set \citep{bib41}. Figure \ref{fig: Model_architecture} displays the overall architecture of our proposed model. In Table~\ref{table7}, the number of parameters in our fine-tuned pretrained model is shown. It is evident that the number of trainable parameters is significantly smaller than the total parameters of the original model. This demonstrates that our proposed model is computationally efficient. By freezing the initial layers, we avoided training the entire model and instead trained only the added layers while leveraging the pretrained weights. This approach not only improved the performance of the pretrained models but also made the model lightweight. 

\textbf{RQ2 Summary}: 
The transfer learning technique has been applied to reduce the complexity of the pretrained CNN model. By freezing the initial layers of an already trained model, additional layers are added for the classification of monkeypox from skin lesion images. This approach reduces the number of trainable parameters in the model, as the weights of the frozen layers are not updated, leading to lower model complexity and faster training time. For example, in the case of a binary dataset, the total parameters of VGG16 are approximately 21 million; however, only 6.4 million are used as trainable parameters.
\subsubsection{Hyperparameters}  
In this study, we utilized a range of hyperparameters to optimize the performance of the deep learning models -VGG16, VGG19, Inception V3, and MobileNetV2- on the monkeypox detection task. These hyperparameters were carefully selected and tuned to balance model accuracy, generalization, and computational efficiency. Below is the description of the key hyperparameters and their significance of our experiments and Table \ref{table6} shows the key values of hyperparameter setting. \\
\textbf{Optimizer}: The optimization algorithm plays a critical role in updating the model wights during training \citep{bib42}. We employed the Adam optimizer due to its adaptive learning rate capabilities and robust performance in complex neural network architectures. The Adam optimizer combines the benefits of momentum and RMSprop, making it suitable for models with varying gradients across layers \citep{bib42}. 
\begin{table}[h!]
\centering
\caption{Hyperparameter settings used for model training.}
\begin{tabular}{|l|c|}
\hline
\textbf{Hyperparameter} & \textbf{Value} \\
\hline
Optimizer   & Adam   \\
Epochs      & 30     \\
Batch Size  & 16     \\
Learning Rate & 0.0001 \\
\hline
\end{tabular}
\label{table6}

\end{table}

\begin{table}[h!]
\centering
\caption{Comparison of trainable and non-trainable parameters for Binary and Multi-Class datasets.}
\begin{tabular}{|l|p{1.7cm}|p{1.7cm}|p{1.7cm}|p{1.7cm}|p{1.7cm}|p{1.7cm}|}
\hline
\multirow{2}{*}{\textbf{Model-Name}} & \multicolumn{3}{c|}{\textbf{Binary Dataset}} & \multicolumn{3}{c|}{\textbf{Multi-Class Dataset}} \\
\cline{2-7}
 & \textbf{Trainable Parameters} & \textbf{Non-trainable Parameters} & \textbf{Total} & \textbf{Trainable Parameters} & \textbf{Non-trainable Parameters} & \textbf{Total} \\
\hline
VGG16 & 6,423,298 (24.50 MB) & 14,714,688 (56.13 MB) & 21,137,986 (80.64 MB) & 6,424,326 (24.51 MB) & 14,714,688 (56.13 MB) & 21,139,014 (80.64 MB) \\
\hline
VGG19 & 6,423,298 (24.50 MB) & 20,024,384 (76.39 MB) & 26,447,682 (100.89 MB) & 6,424,326 (24.51 MB) & 20,024,384 (76.39 MB) & 26,448,710 (100.89 MB) \\
\hline
InceptionV3 & 13,107,970 (50.00 MB) & 21,802,784 (83.17 MB) & 34,910,754 (133.17 MB) & 13,108,998 (50.01 MB) & 21,802,784 (83.17 MB) & 34,911,782 (133.18 MB) \\
\hline
MobileNetV2 & 16,057,090 (61.25 MB) & 2,257,984 (8.61 MB) & 18,315,074 (69.87 MB) & 16,058,118 (61.26 MB) & 2,257,984 (8.61 MB) & 18,316,102 (69.87 MB) \\
\hline
\end{tabular}

\label{table7}
\end{table}

\textbf{Loss Function}: For our classification task, we utilized the categorical cross-entropy loss function . It is widely used for multi-class classification problems. This function computes the logarithmic difference between the true labels and the predicted probabilities.\\ 
\textbf{Epochs}: The number of epochs was set to a maximum of 30 to allow sufficient training while mitigating the risk of overfitting. To prevent excessive training, we employed an early stopping mechanism as described below.\\ 
\textbf{Batch Size}: The batch size was set to 16, for balancing computational efficiency and model performance.  This size was chosen to ensure stable gradient updates while fitting within the memory constraints of the training environment .\\
\textbf{Early stopping}: Early stopping was implemented to monitor the validation loss and stop training if no improvement was observed for consecutive epochs. This mechanism helped in preventing overfitting and reducing unnecessary computational overhead.\\ 
\textbf{Model Checkpoint save}: To retain the best performing model during training, we utilized a model check point strategy. The model 
Learning rate: The learning rate was initialized at 0.001, providing a moderate step size for weight updates. A constant learning rate was used for simplicity. It worked well in conjunction with the Adam optimizer. \\
\textbf{Patience}: The patience parameter for early stopping set to 5, allowing the model adequate epochs to recover from minor fluctuations in validation performance before stopping the training process. \\
\textbf{RQ3 Summary}: For the hyperparameter settings, model checkpoint saving, and early stopping have been applied. These two steps are crucial for training the model efficiently and avoiding unnecessary training, thereby saving time and computational resources. Checkpoint saving helps resume training from the last saved state if interrupted, while early stopping halts training after consecutive steps when the model's performance starts to decline.
\subsubsection{Evaluation techniques}  
To measure the performance of each pre-trained model the following metrics are considered \citep{bib40}. \\
\textbf{Accuracy}: Accuracy is the ratio of correctly predicted observations (both positive and negative) to the total number of observations. It measures the overall correctness of a model. 
\begin{equation}
    \text{Accuracy} = \frac{TP + FP}{\text{Total}}
 \end{equation}
\textbf{Precision}: Precision is the ratio of correctly predicted positive observations to the total predicted positives. It indicates how many of the predicted positive instances were correct.
\begin{equation}
    \text{Precision} = \frac{TP}{\text{TP+FP}}
 \end{equation}
\textbf{Recall}: Recall is the ratio of correctly predicted positive observations to all actual positive observations. It measures the ability of a model to capture all relevant instances.
\begin{equation}
    \text{Recall} = \frac{TP}{\text{TP+FN}}
 \end{equation}
\textbf{F1-score}: F1-score is the harmonic mean of Precision and Recall. It balances the trade-off between precision and recall, especially in cases of class imbalance.
\begin{equation}
    F1\text{-}score = \frac{2 \times \text{Precision} \times \text{Recall}}{\text{Precision} + \text{Recall}}
 \end{equation}
\\

\textbf{Wall time}: Wall time refers to the actual elapsed time taken for a process to complete, as experienced in real time (including time spent on waiting, I/O operations, etc.).\\
\textbf{CPU time}: CPU time refers to the actual time the CPU spends computing for a specific process. It excludes time spent on waiting for I/O or other system resources.\\
\textbf{ROC/AUC (Receiver Operating Characteristic / Area Under Curve)}: It is A graphical representation of a model’s performance by plotting the True Positive Rate (Recall) against the False Positive Rate (FPR) at various classification thresholds.\\
AUC is a single scalar value representing the area under the ROC curve. AUC ranges from 0 to 1: 1 means perfect model. 0 means completely incorrect prediction.Here, TP=True Positive (Actual and predicted both positive), FP = False Positive (Actual Positive but predicted negative), TN= True Negative (Actual and predicted both negative), FN= False Negative (Actual negative but predicted positive) 

\section{Experiment Results}
\textbf{RQ1}: Which pre-trained CNN models (VGG16, VGG19, InceptionV3, MobileNetV2) demonstrate the best performance in detecting monkeypox from skin lesion images based on key evaluation metrics (accuracy, precision, recall, F1-score, execution time)? \\
\textbf{RQ4}: Which technique can be applied to improve the interpretability of the model’s performance?
In this section these two research questions are answered.
\subsection{Binary Dataset}
Table \ref{table8} provides a comparison of different deep learning models (VGG-16, VGG-19, Inception-V3, MobileNetV2) based on their performance metrics and execution times. 
VGG-16 and VGG-19 show similar results, with their precision, recall, and F1 score mainly 0.89. The F1-score for VGG-19 is slightly better at 0.89, compared to 0.88 for VGG-16, a difference of 0.01. The loss score for both models are 0.27.\\

\begin{figure}[htbp]
    \centering

    \begin{subfigure}[b]{0.45\columnwidth}  
        \centering
        \includegraphics[width=\linewidth]{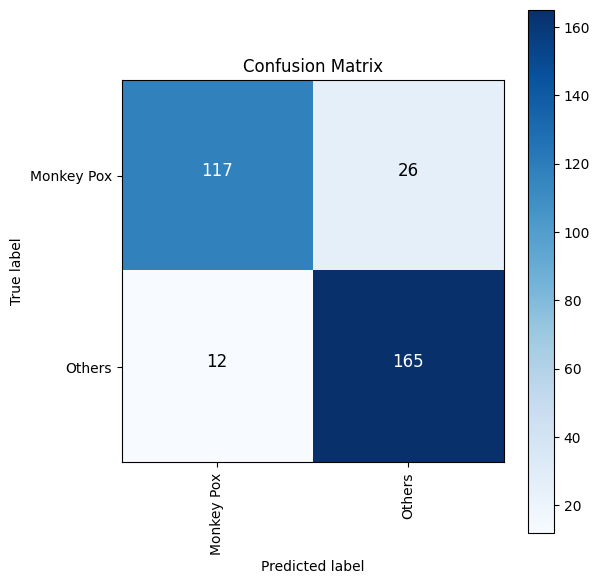}
        \caption{VGG-16}
        \label{fig:cmb1}
    \end{subfigure}
    \hfill
    \begin{subfigure}[b]{0.45\columnwidth}  
        \centering
        \includegraphics[width=\linewidth]{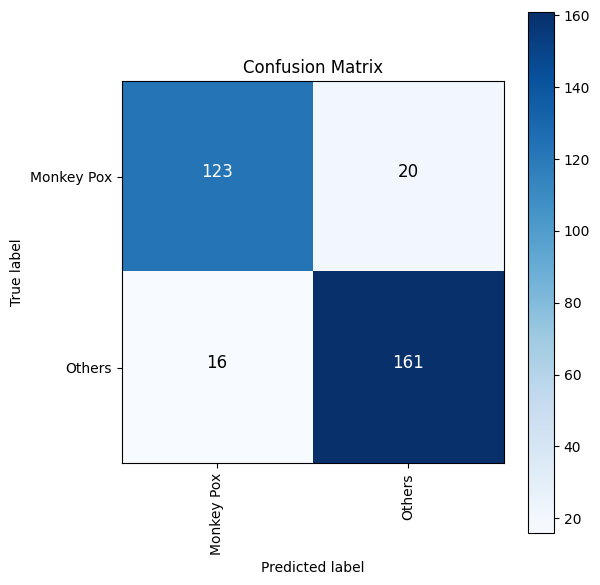}
        \caption{VGG19}
        \label{fig:cmb2}
    \end{subfigure}

    \vspace{0.5em}

    \begin{subfigure}[b]{0.45\columnwidth}  
        \centering
        \includegraphics[width=\linewidth]{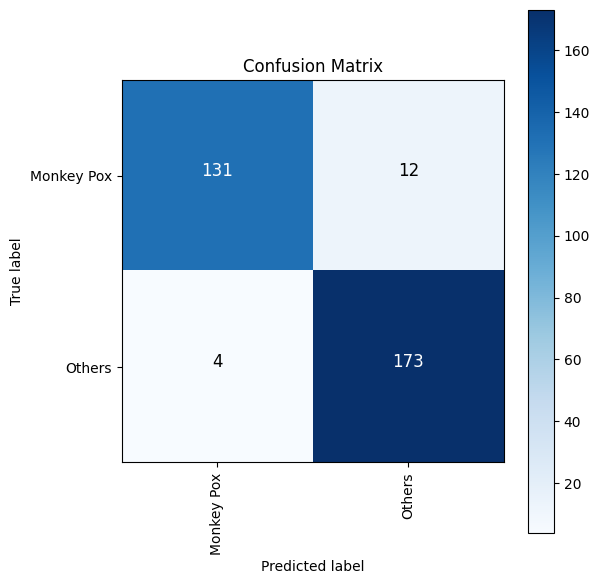}
        \caption{InceptionV3}
        \label{fig:cmb3}
    \end{subfigure}
    \hfill
    \begin{subfigure}[b]{0.45\columnwidth}  
        \centering
        \includegraphics[width=\linewidth]{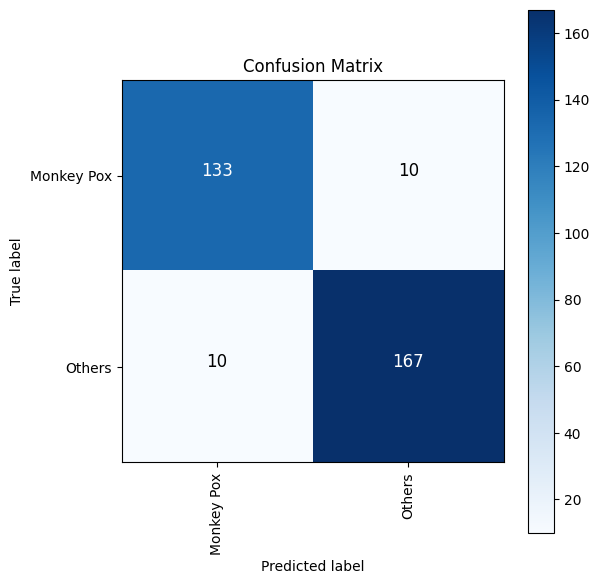}
        \caption{MobileNetV2}
        \label{fig:cmb4}
    \end{subfigure}

    \caption{Confusion matrices for four different models for binary }
    \label{fig:all_conf_matrices_bin}
\end{figure}

\textbf{Analysis of confusion matrices}:
 \textbf{Figure \ref{fig:all_conf_matrices_bin}} shows the confusion matrix of all four models.As shown in Figure \ref{fig:cmb1} and \ref{fig:cmb2}, VGG-16 makes 38 incorrect predictions, while VGG-19 makes 36 incorrect predictions among the test data. The wall times for both models are also similar, approximately 22 minutes, as shown in Table \ref{table8}. However, the CPU time is slightly higher for VGG-19, at 5 minutes 41 seconds, compared to 3 minutes 43 seconds for VGG-16. Overall, their performance in terms of accuracy and other metrics is nearly identical.
Inception-V3 achieves accuracy, precision, recall, and F1-score of 0.95, outperforming the VGG models. This model correctly predicts 304 cases while making 16 incorrect predictions, as shown in Figure \ref{fig:cmb3}. Additionally, it has a much lower loss score of 0.11, indicating better generalization and fewer errors. The wall time is reduced to 15 minutes 50 seconds, making it significantly faster than the VGG models despite its superior performance. This efficiency can be attributed to the optimized Inception modules. The CPU time, at 3 minutes 60 seconds, is slightly higher than that of VGG-16 but still more efficient than VGG-19.
MobileNetV2 achieves accuracy, precision, recall, and F1-score of 0.94, which is slightly lower than Inception-V3 but better than the VGG models. Its loss score is 0.17, higher than that of Inception-V3 but still better than the VGG models. Among the test data, MobileNetV2 makes 20 incorrect predictions, as shown in Figure \ref{fig:cmb4}. The wall time is significantly lower, at 6 minutes 48 seconds, making it the fastest among all models. It also has the lowest CPU time, at 3 minutes 16 seconds, reflecting its lightweight architecture optimized for speed.

\begin{table}[h!]
\centering
\caption{Model wise performance analysis for binary classification.}

\resizebox{\columnwidth}{!}{

\begin{tabular}{|l|c|c|c|c|c|p{1.5cm}|p{1.5cm}|}

\hline
\textbf{Model Name} & \textbf{Accuracy} & \textbf{Precision} & \textbf{Recall} & \textbf{F1-score} & \textbf{Loss} & \textbf{Wall Time} & \textbf{CPU Time} \\
\hline
VGG-16 & 0.89 & 0.89 & 0.88 & 0.88 & 0.27 & 22m 8s & 3m 43s \\
\hline
VGG-19 & 0.89 & 0.89 & 0.88 & 0.89 & 0.27 & 22m 53s & 5m 41s \\
\hline
Inception-V3 & 0.95 & 0.95 & 0.95 & 0.95 & 0.11 & 15m 50s & 3m 60s \\
\hline
MobileNetV2 & 0.94 & 0.94 & 0.94 & 0.94 & 0.17 & 6m 48s & 3m 16s \\
\hline
\end{tabular}
}
\label{table8}
\end{table}

\begin{figure}[htbp]
    \centering

        \begin{subfigure}[b]{0.45\textwidth}
            \includegraphics[width=\linewidth]{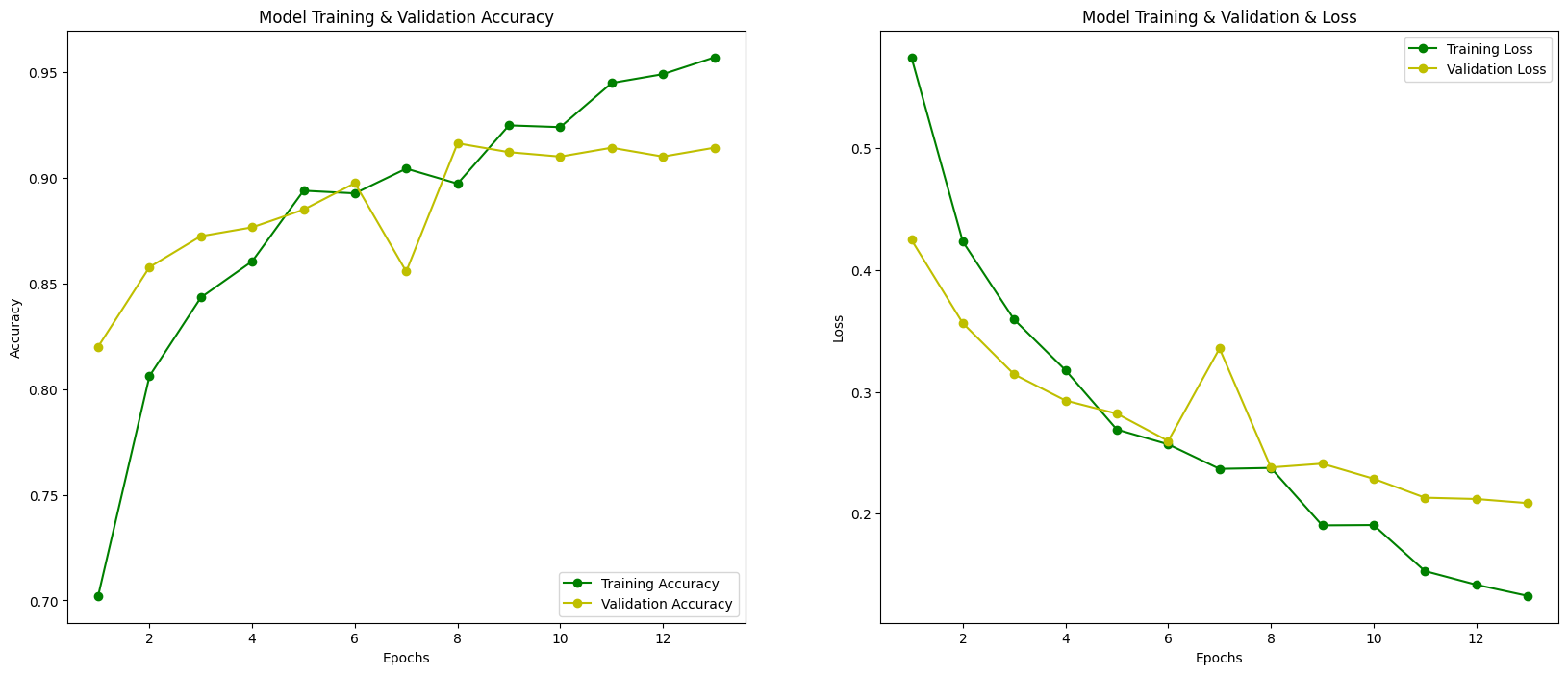}
            \caption{VGG16}
            \label{vgg16_bin_acc}
        \end{subfigure}
        \hspace{1em}
        \begin{subfigure}[b]{0.45\textwidth}
            \includegraphics[width=\linewidth]{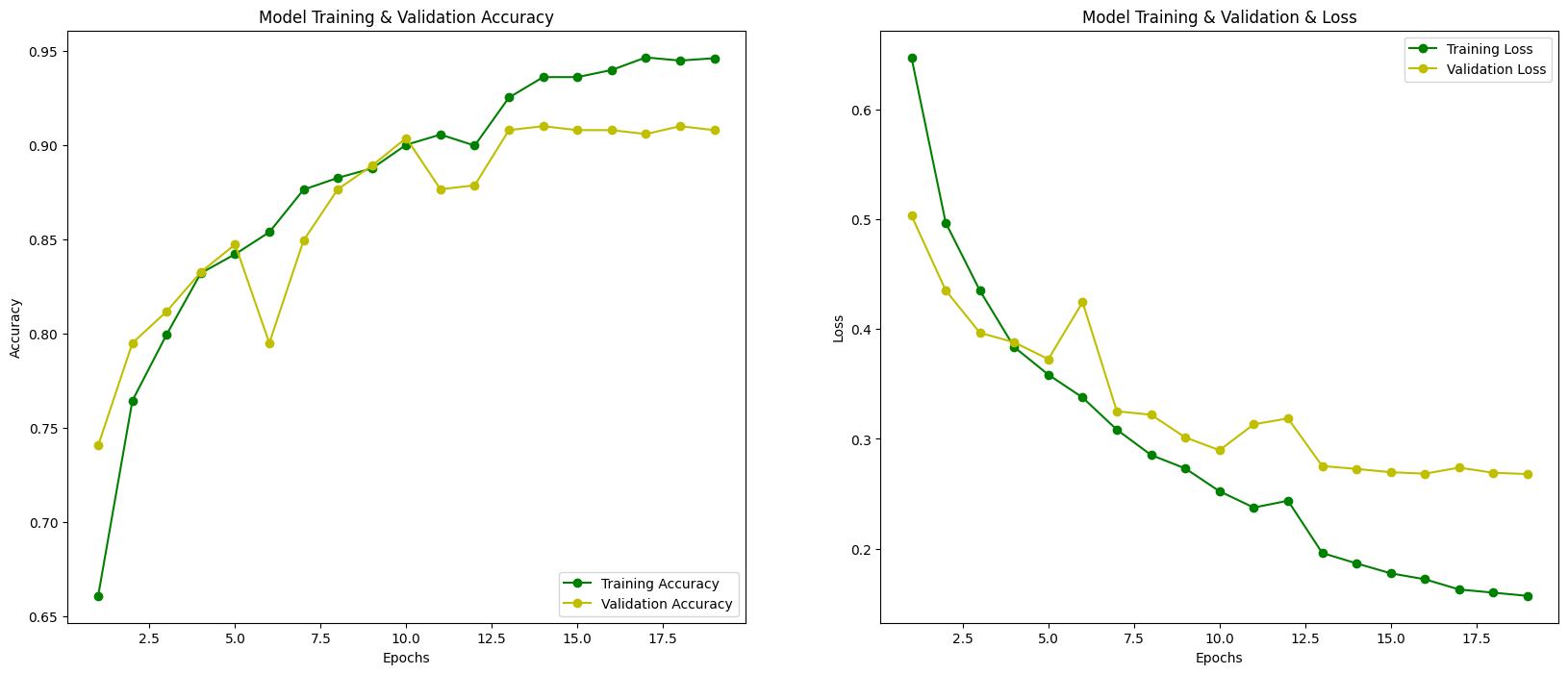}
            \caption{VGG19}
             \label{vgg19_bin_acc}
        \end{subfigure}

        \vspace{1em}

        \begin{subfigure}[b]{0.45\textwidth}
            \includegraphics[width=\linewidth]{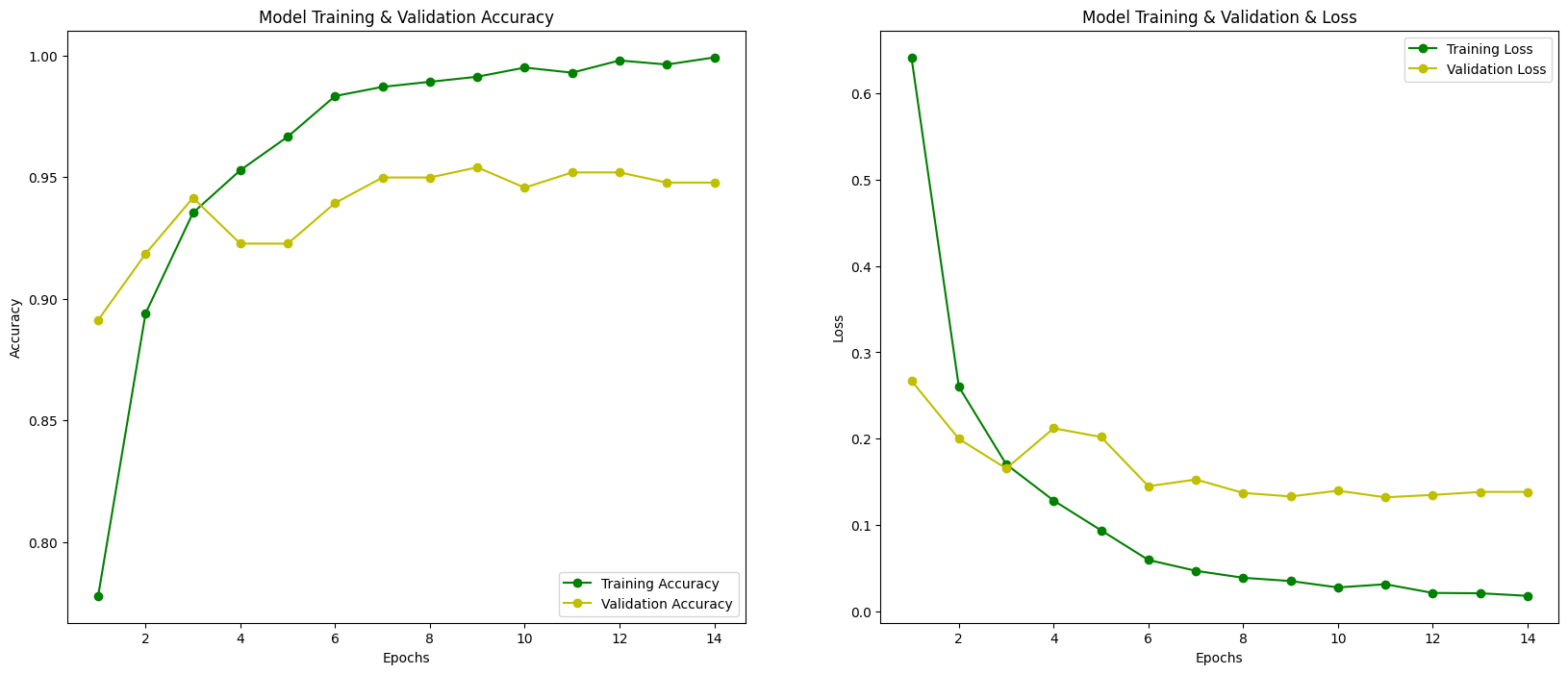}
            \caption{Inception}
             \label{incep_bin_acc}
        \end{subfigure}
        \hspace{1em}
        \begin{subfigure}[b]{0.45\textwidth}
            \includegraphics[width=\linewidth]{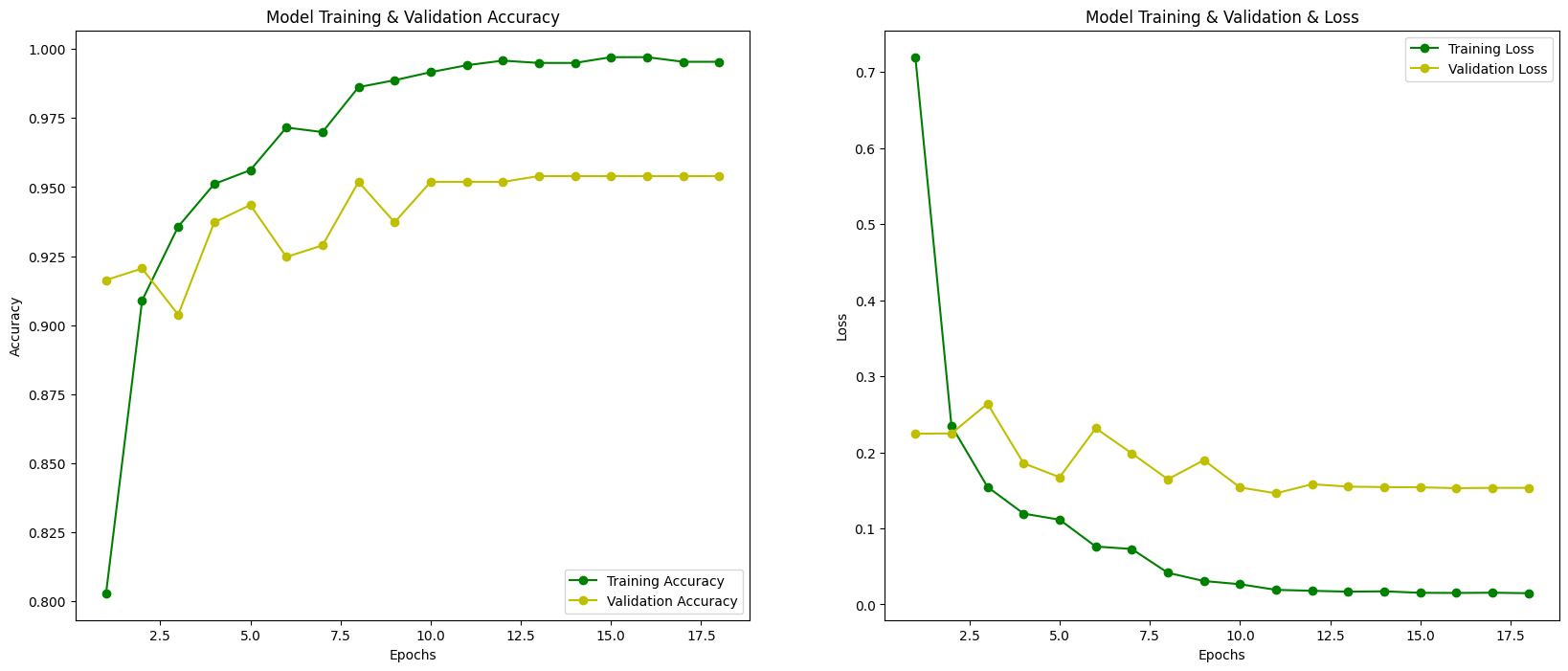}
            \caption{MobileNetV2}
             \label{mobile_bin_acc}
        \end{subfigure}

    \caption{Accuracy-loss curve for four different models for binary class}
    \label{fig:all_acc_loss_bin}
\end{figure}

\begin{figure*}[htbp]
    \centering
 
    \begin{subfigure}[b]{\textwidth}
        \centering

        \begin{subfigure}[b]{0.45\textwidth}
            \includegraphics[width=\linewidth]{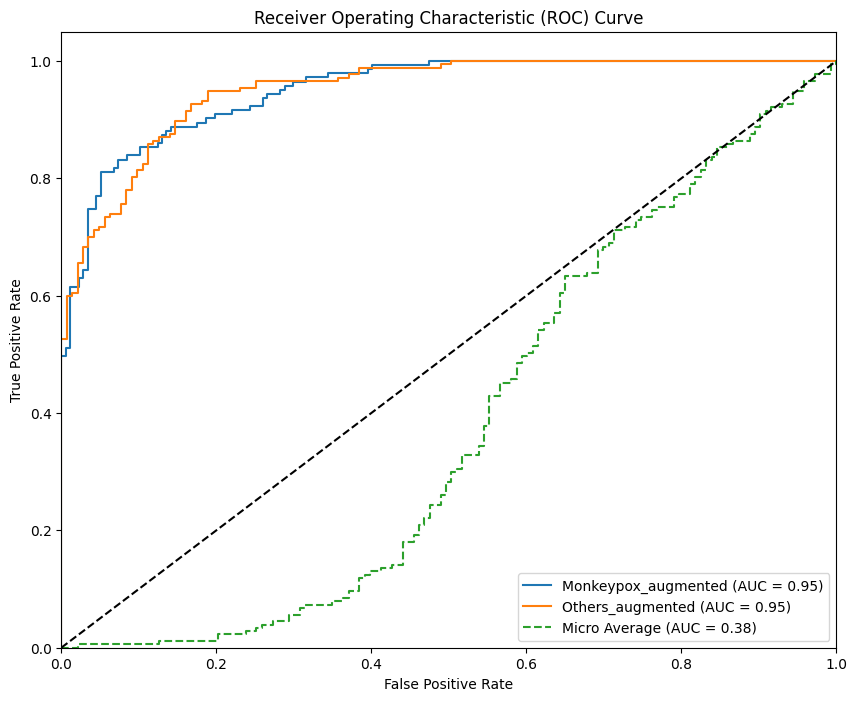}
            \caption{VGG16}
            \label{vgg16_bin_roc}
        \end{subfigure}
        \hspace{1em}
        \begin{subfigure}[b]{0.45\textwidth}
            \includegraphics[width=\linewidth]{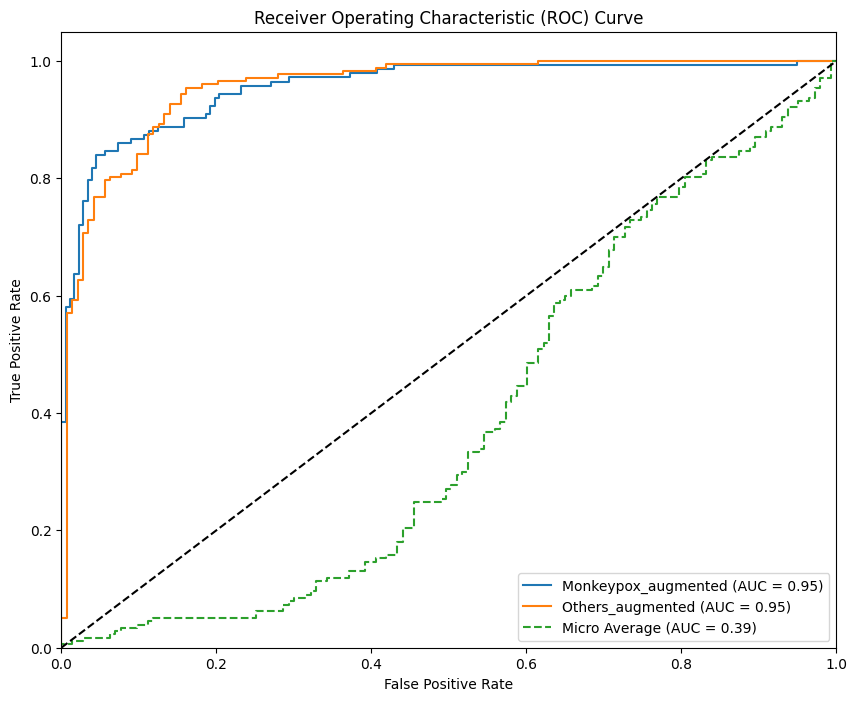}
            \caption{VGG19}
             \label{vgg19_bin_roc}
        \end{subfigure}

        \vspace{1em}

        \begin{subfigure}[b]{0.45\textwidth}
            \includegraphics[width=\linewidth]{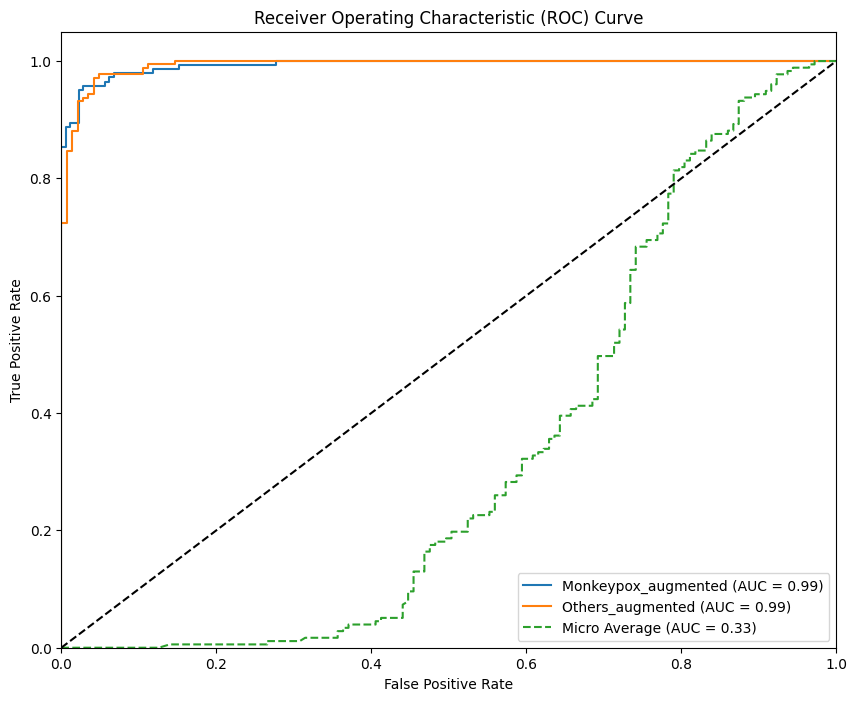}
            \caption{Inception}
             \label{incep_bin_roc}
        \end{subfigure}
        \hspace{1em}
        \begin{subfigure}[b]{0.45\textwidth}
            \includegraphics[width=\linewidth]{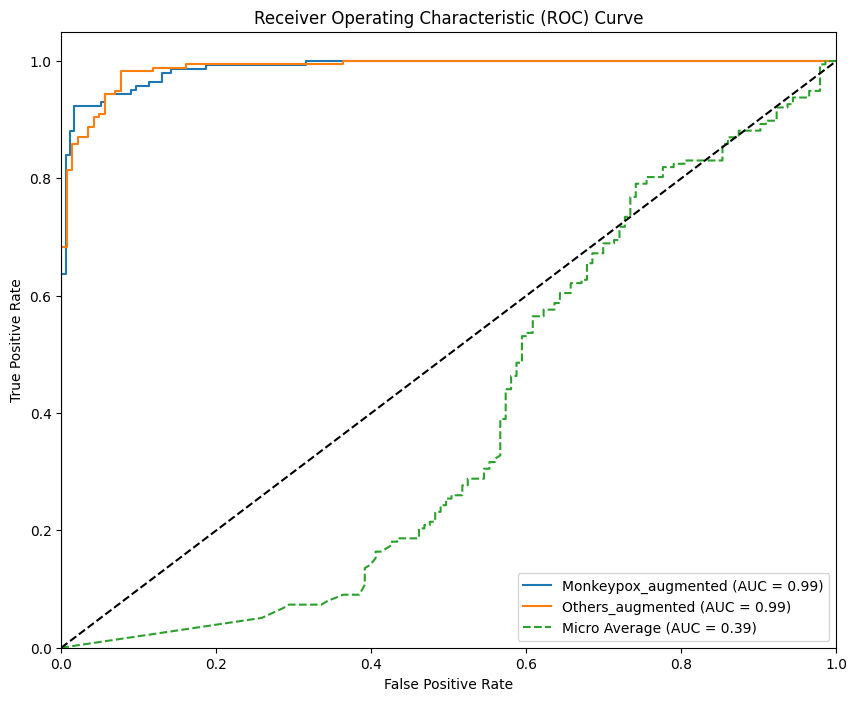}
            \caption{MobileNetV2}
             \label{mobile_bin_roc}
        \end{subfigure}

    \end{subfigure}

    \vspace{2em}

   
    \caption{Visual comparison of model performance: ROC curves}
    \label{fig:roc_all_bin}
\end{figure*}

\begin{figure*}[t]
    \centering

    \begin{subfigure}[b]{0.45\textwidth}
        \centering
        \includegraphics[width=\linewidth]{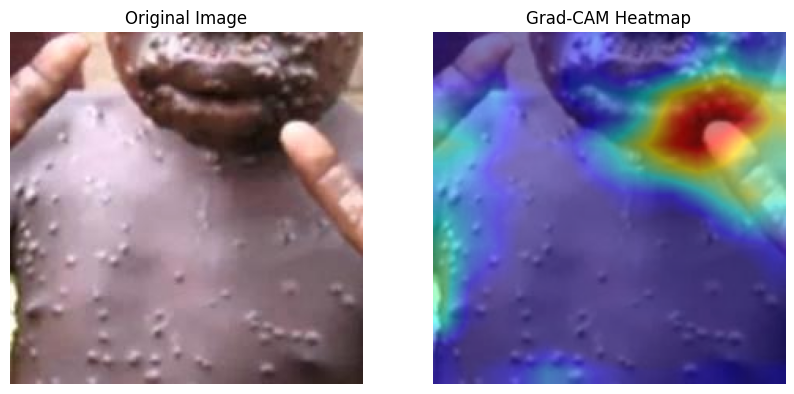}
        \caption{VGG16 Grad-CAM}
        \label{fig:gradcam_vgg16}
    \end{subfigure}
    \hfill
    \begin{subfigure}[b]{0.45\textwidth}
        \centering
        \includegraphics[width=\linewidth]{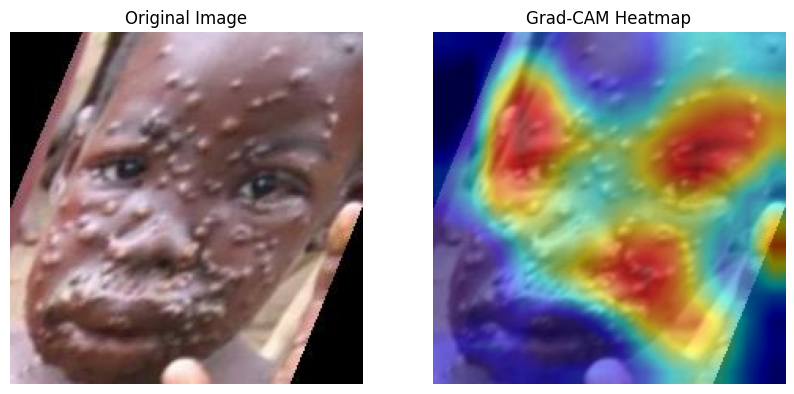}
        \caption{VGG19 Grad-CAM}
        \label{fig:gradcam_vgg19}
    \end{subfigure}

    \vspace{0.5em}

    \begin{subfigure}[b]{0.45\textwidth}
        \centering
        \includegraphics[width=\linewidth]{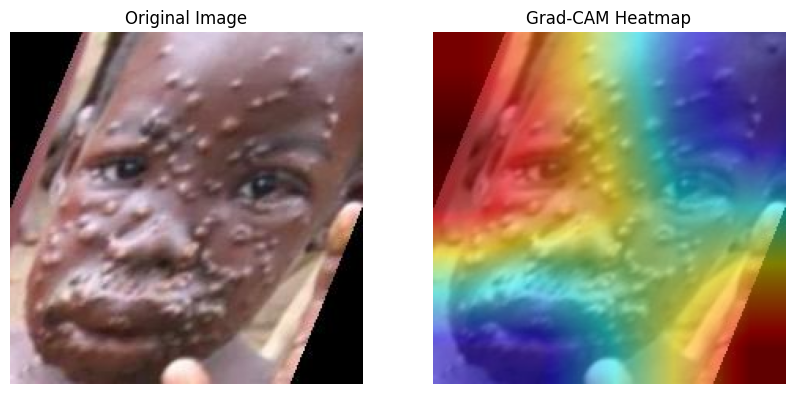}
        \caption{Inception Grad-CAM}
        \label{fig:gradcam_incep}
    \end{subfigure}
    \hfill
    \begin{subfigure}[b]{0.45\textwidth}
        \centering
        \includegraphics[width=\linewidth]{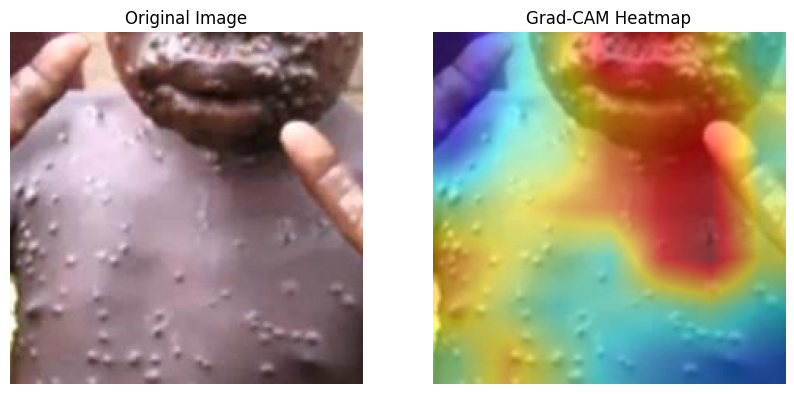}
        \caption{MobileNetV2 Grad-CAM}
        \label{fig:gradcam_mobilenet}
    \end{subfigure}

    \caption{Grad-CAM visualizations for the four models for binary class.}
    \label{fig:gradcam_all_bin}
\end{figure*}

\textbf{Analysis of accuracy and loss curves: }
Analysis of \textbf{Figure \ref{fig:all_acc_loss_bin}}:
Figure \ref{vgg16_bin_acc} for VGG16 model, training accuracy increases steadily, reaching around 0.95 by the end. Validation accuracy also rises to approximately 0.90, showing a similar trend with some fluctuation around epoch 8. For loss curve both training and validation loss decrease consistently, with training loss reaching a minimum around 0.1. Validation loss fluctuates more, indicating some instability, possibly due to early signs of overfitting.
Figure \ref{vgg19_bin_acc} for VGG19 model, the training accuracy improves continuously, stabilizing around 0.91. Validation accuracy follows a similar pattern, converging around 0.88 with a small drop around epoch 7, suggesting it matches the training accuracy but with slightly more variation. For loss curve, both training and validation losses decrease smoothly, indicating good model performance and stable learning. The training loss falls below 0.2 by the last epoch, while the validation loss plateaus around 0.3, suggesting the model generalizes well with minimal overfitting.
In Figure \ref{incep_bin_acc} accuracy curve of InceptionV3, the training accuracy reaches close to 1.0 by the final epochs, suggesting the model has learned the training data almost completely. Validation accuracy stabilizes around 0.92, indicating good generalization. For loss curve in training loss reduces to near-zero, while validation loss levels off around 0.1. The trend suggests that the model achieves excellent performance on the training data, with some risk of slight overfitting due to the high accuracy disparity between training and validation by the final epochs.
In Figure \ref{mobile_bin_acc} accuracy and loss curve for MobileNetV2, the training loss steadily decreases throughout training, falling below 0.1 by the final epoch. This smooth downward trend indicates effective error minimization on the training data. In contrast, the validation loss shows a rapid decline initially but then plateaus around 0.2-0.3, suggesting that while the model generalizes reasonably well, its performance on the validation set does not improve further after a certain point. The use of early stopping effectively prevents overfitting, preserving the model’s generalization capability.
Overall, there are no signs of underfitting in these graphs, as the models reach high accuracy for both training and validation datasets across all models. In the third and fourth model, the near-zero training loss alongside a higher validation loss suggests some degree of overfitting, as the model has memorized the training data well but generalizes slightly less to unseen data. Early stopping was applied, which halted training before the 30 epochs set initially. This prevented unnecessary overfitting and saved computation time. For instance, in Model 1, training stopped at epoch 12, where both accuracy and loss reached stable values, indicating optimal performance without significant overfitting.

\textbf{Analysis of roc curves: }
\textbf{Figure \ref{fig:roc_all_bin}} displays the roc curves of all models.
The roc curves  of four deep learning models for monkeypox detection are compared in Figure \ref{vgg16_bin_roc},Figure \ref{vgg19_bin_roc}, Figure \ref{incep_bin_roc} and Figure \ref{mobile_bin_roc} respectively. While Inception V3 and MobileNetV2 have a greater class-specific AUC of 0.99, suggesting near-perfect performance in class distinction, VGG16 and VGG19 both obtain an AUC of 0.95 for monkeypox and other classes. Nonetheless, all models' micro-average AUC values, which represent overall classification, are still somewhat low (0.38 for VGG16, 0.39 for VGG19 and MobileNetV2, and 0.33 for Inception V3), indicating difficulties managing class imbalance or optimizing general classification. Class-specific discrimination is where Inception V3 and MobileNetV2 shine, but overall classification performance is still lacking.

\textbf{Feature highlights using Explainable AI techniques (Grad-CAM) for binary class data}

In \textbf{Figure \ref{fig:gradcam_all_bin}}, output heatmap generated by Gradient-weighted Class Activation Mapping (GRAD-CAM) \citep{bib43} is shown. This technique utilizes the gradients of a target concept (such as 'dog' in a classification network or a sequence of words in a captioning network) that flow into the final convolutional layer. It generates a coarse localization map, emphasizing the critical regions in the image that contribute to predicting the concept \citep{bib43}. From the following figure, the highlighted region with dark portions gives the pixel information that helps to identify the portion that mainly contributes for the detection of monkeypox virus. Dark red regions indicate areas of high importance, meaning these areas contributed the most to the model's prediction for the target concept. Blue (or cooler colors) regions represent areas of low importance, meaning these areas had little to no contribution to the prediction. This technique enhances model interpretability, allowing us to assess the trustworthiness of our pretrained CNN model's performance. It aids healthcare practitioners in making informed decisions about diseases while incorporating manual intervention when necessary.

\subsection{Multi-Class Dataset}
\begin{figure}[htbp]
    \centering

    \begin{subfigure}[b]{0.45\columnwidth}  
        \centering
        \includegraphics[width=\linewidth]{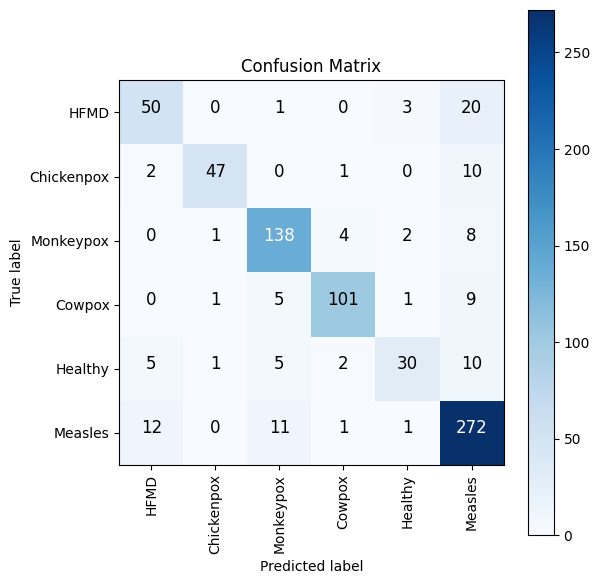}
        \caption{VGG-16}
        \label{fig:cm1}
    \end{subfigure}
    \hfill
    \begin{subfigure}[b]{0.45\columnwidth}  
        \centering
        \includegraphics[width=\linewidth]{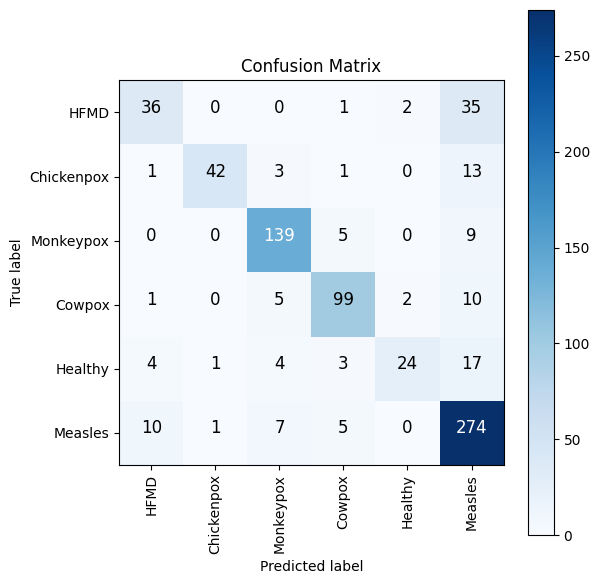}
        \caption{VGG19}
        \label{fig:cm2}
    \end{subfigure}

    \vspace{0.5em}

    \begin{subfigure}[b]{0.45\columnwidth}  
        \centering
        \includegraphics[width=\linewidth]{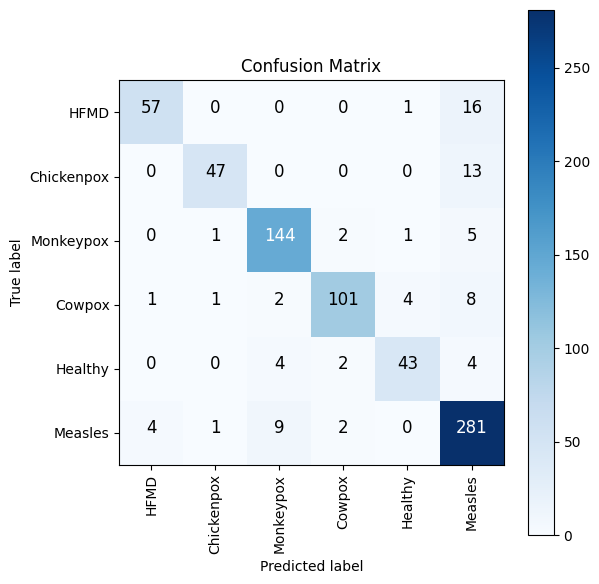}
        \caption{InceptionV3}
        \label{fig:cm3}
    \end{subfigure}
    \hfill
    \begin{subfigure}[b]{0.45\columnwidth}  
        \centering
        \includegraphics[width=\linewidth]{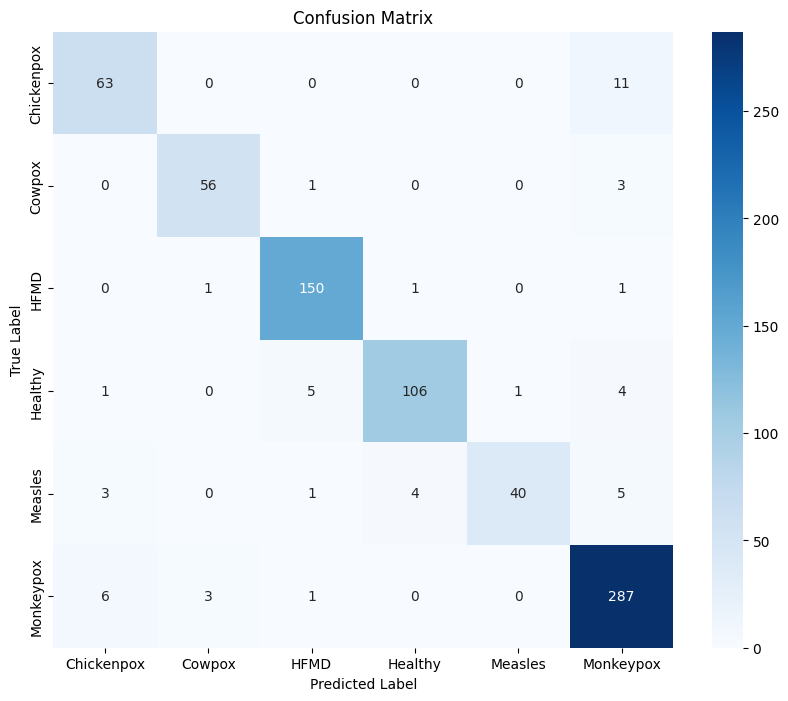}
        \caption{MobileNetV2}
        \label{fig:cm4}
    \end{subfigure}

    \caption{Confusion matrices for four different models for multi-class dataset }
    \label{fig:confusion_mul}
\end{figure}

From \textbf{Table \ref{table9}} it can be seen that VGG-16 and VGG-19 demonstrate similar performance, with VGG-16 achieving accuracy, precision, recall, and F1-score of 0.85, 0.85, 0.78, and 0.81, respectively, while VGG-19 achieves 0.81, 0.84, 0.72, and 0.76. The F1-score for VGG-16 is better by 0.05 compared to VGG-19. Loss scores are 0.41 for VGG-16 and 0.54 for VGG-19, indicating better generalization for VGG-16.\\
\textbf{Analysis of Confusion Matrices:}
 \textbf{Figure \ref{fig:confusion_mul}} displays the confusion matrices for all four models for multi class dataset. it is evident, VGG-16 makes fewer incorrect predictions overall. The wall times for both models are similar, at approximately 1 hour 44 minutes for VGG-16 and 1 hour 39 minutes for VGG-19. However, the CPU time is higher for VGG-19 at 17 minutes 48 seconds compared to 14 minutes 54 seconds for VGG-16, reflecting its deeper architecture. Overall, VGG-16 outperforms VGG-19 in both accuracy and computational efficiency.

\begin{table}[h!]
\centering
\caption{Model wise performance analysis for multi-class classification.}

\resizebox{\columnwidth}{!}{
\begin{tabular}{|l|c|c|c|c|c|p{1.5cm}|p{1.5cm}|}
\hline
\textbf{Model Name} & \textbf{Accuracy} & \textbf{Precision} & \textbf{Recall} & \textbf{F1-score} & \textbf{Loss} & \textbf{Wall Time} & \textbf{CPU Time} \\
\hline
VGG-16 & 0.85 & 0.85 & 0.78 & 0.81 & 0.41 & 1h 44m & 14m 54s \\
\hline
VGG-19 & 0.81 & 0.84 & 0.72 & 0.76 & 0.54 & 1h 39m & 17m 48s \\
\hline
Inception-V3 & 0.89 & 0.91 & 0.85 & 0.88 & 0.55 & 1h 44m & 14m 54s \\
\hline
MobileNetV2 & 0.93 & 0.93 & 0.90 & 0.91 & 0.32 & 56m 58s & 6m 9s \\
\hline
\end{tabular}
}
\label{table9}
\end{table}

\begin{figure}[htbp]
    \centering

        \begin{subfigure}[b]{0.45\textwidth}
            \includegraphics[width=\linewidth]{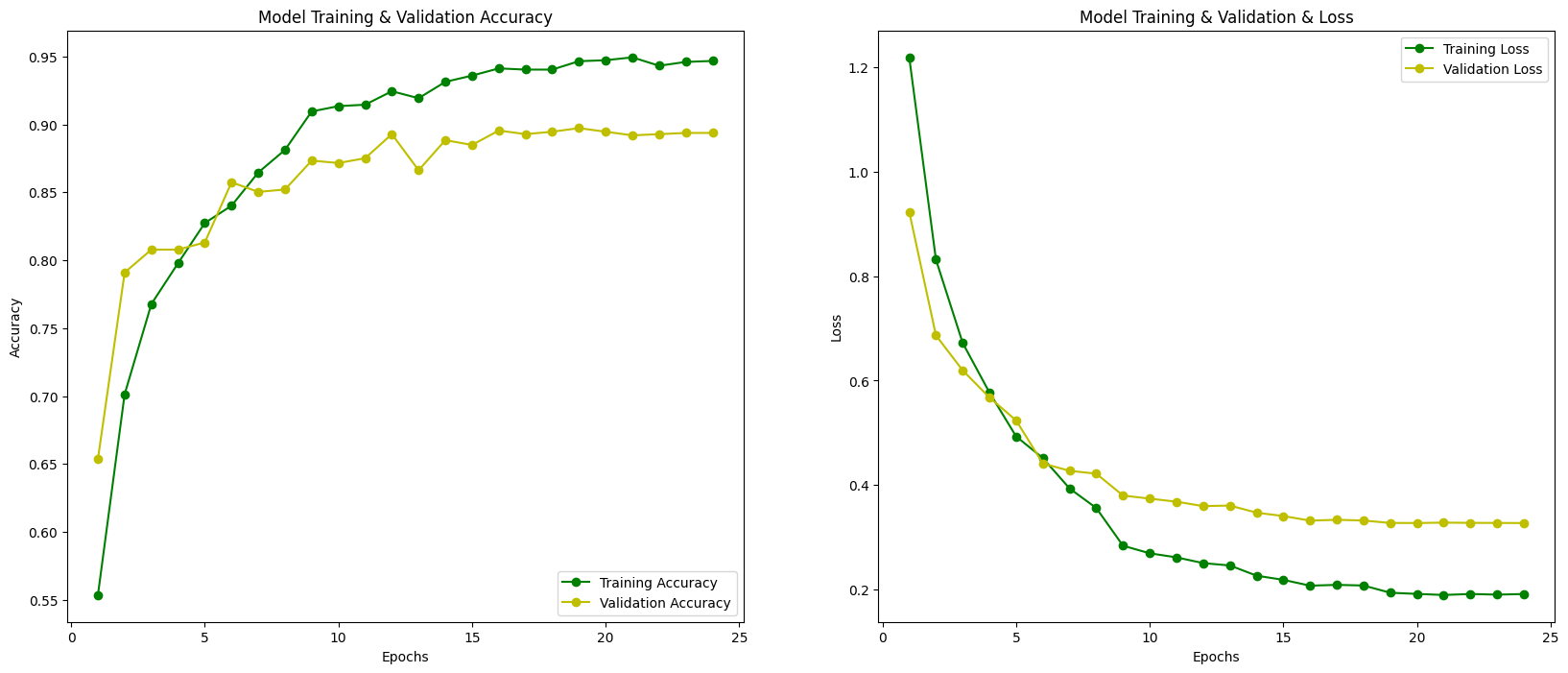}
            \caption{VGG16}
            \label{vgg16_mul_acc}
        \end{subfigure}
        \hspace{1em}
        \begin{subfigure}[b]{0.45\textwidth}
            \includegraphics[width=\linewidth]{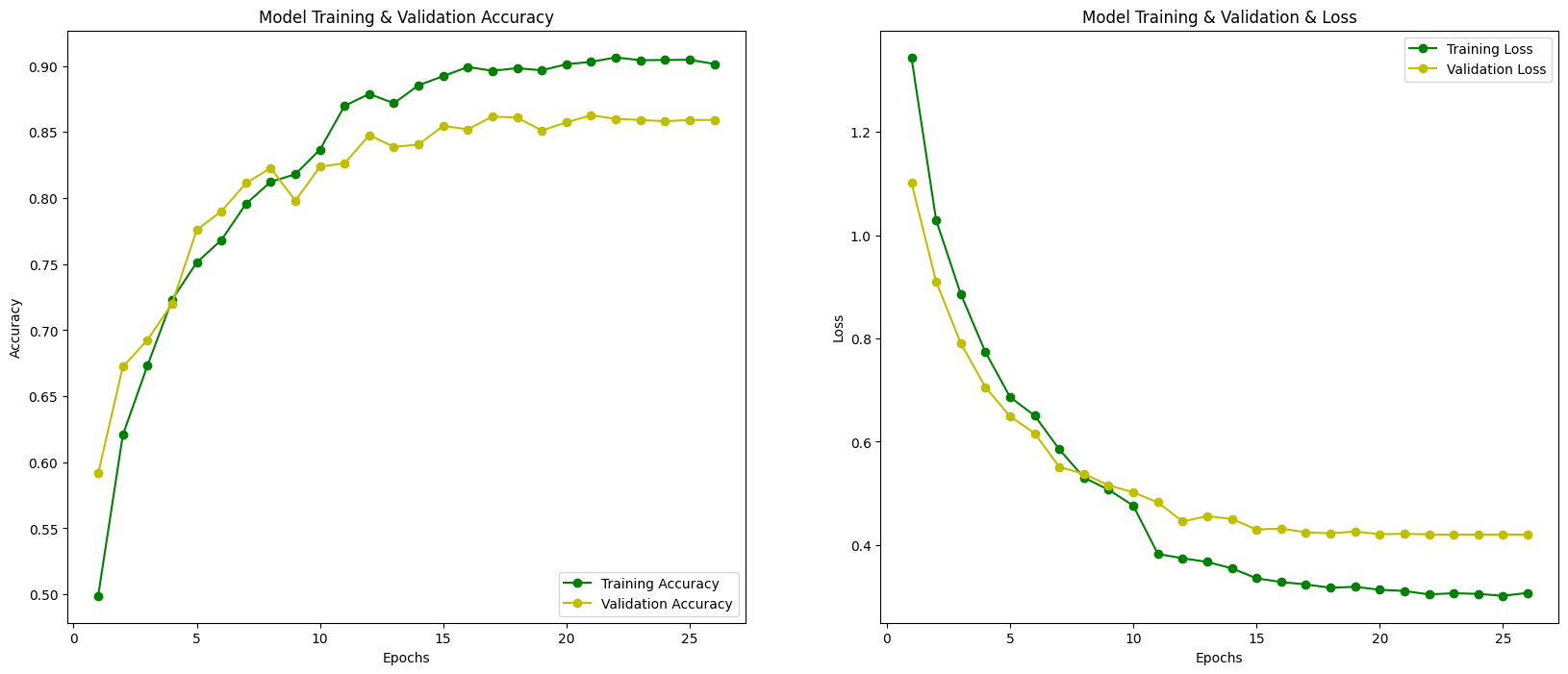}
            \caption{VGG19}
            \label{vgg19_mul_acc}
        \end{subfigure}

        \vspace{1em}

        \begin{subfigure}[b]{0.45\textwidth}
            \includegraphics[width=\linewidth]{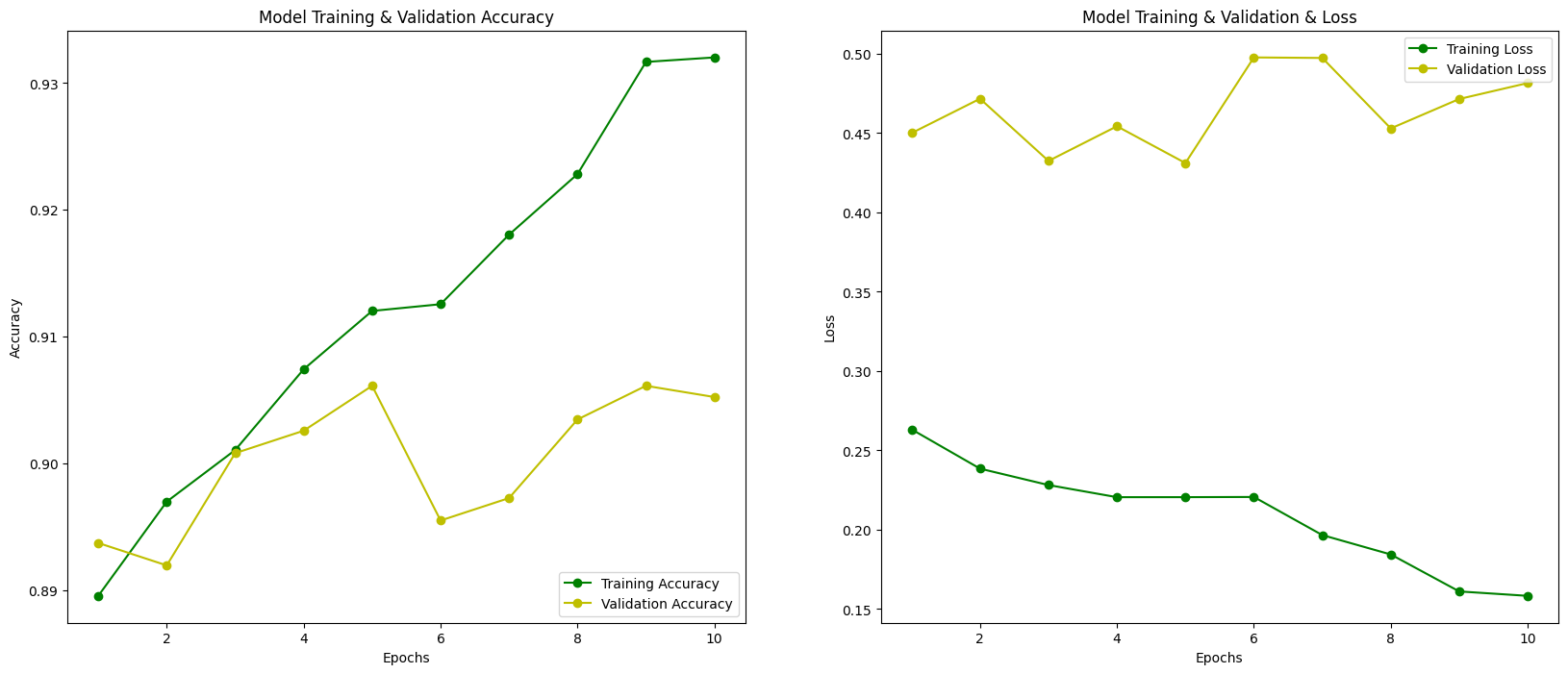}
            \caption{Inception}
            \label{incep_mul_acc}
        \end{subfigure}
        \hspace{1em}
        \begin{subfigure}[b]{0.45\textwidth}
            \includegraphics[width=\linewidth]{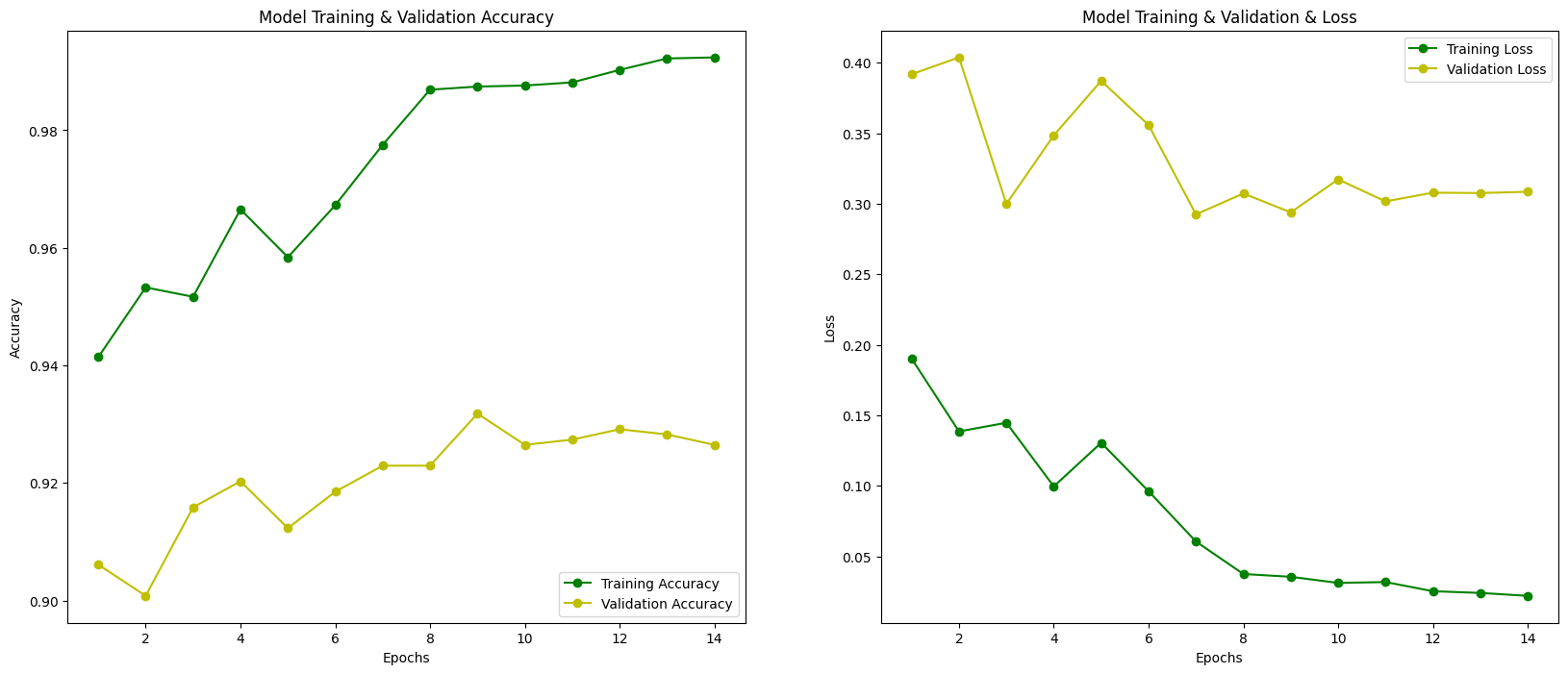}
            \caption{MobileNetV2}
            \label{mobile_mul_acc}
        \end{subfigure}

        \caption{ Accuracy and loss curves for all the four models for multi-class dataset}
    \label{fig:all_acc_loss_mul}
   
\end{figure}


\begin{figure*}[htbp]
    \centering

        \begin{subfigure}[b]{0.45\textwidth}
            \includegraphics[width=\linewidth]{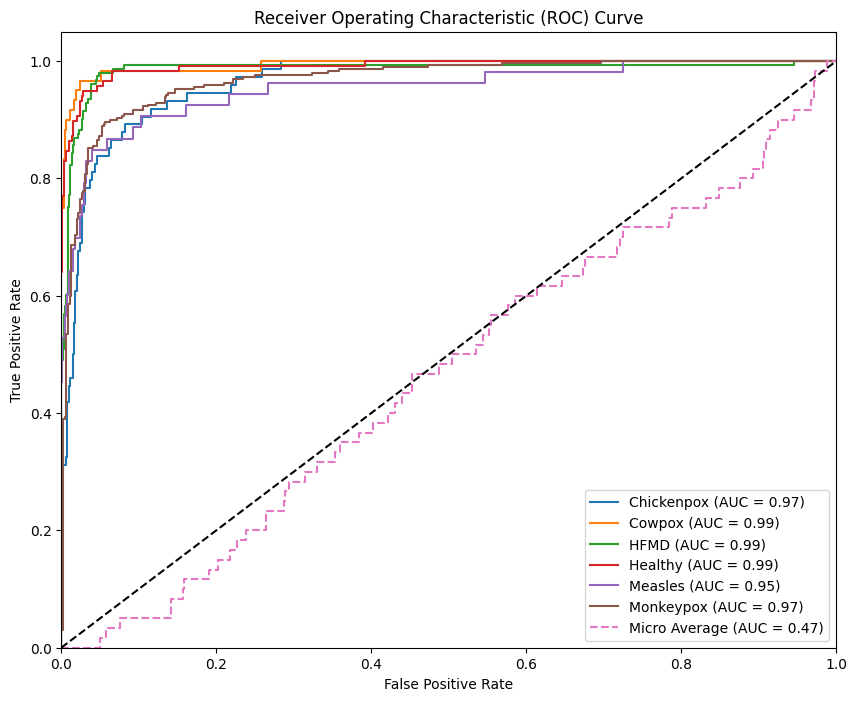}
            \caption{VGG16}
            \label{vgg16_mul_roc}
        \end{subfigure}
        \hspace{1em}
        \begin{subfigure}[b]{0.45\textwidth}
            \includegraphics[width=\linewidth]{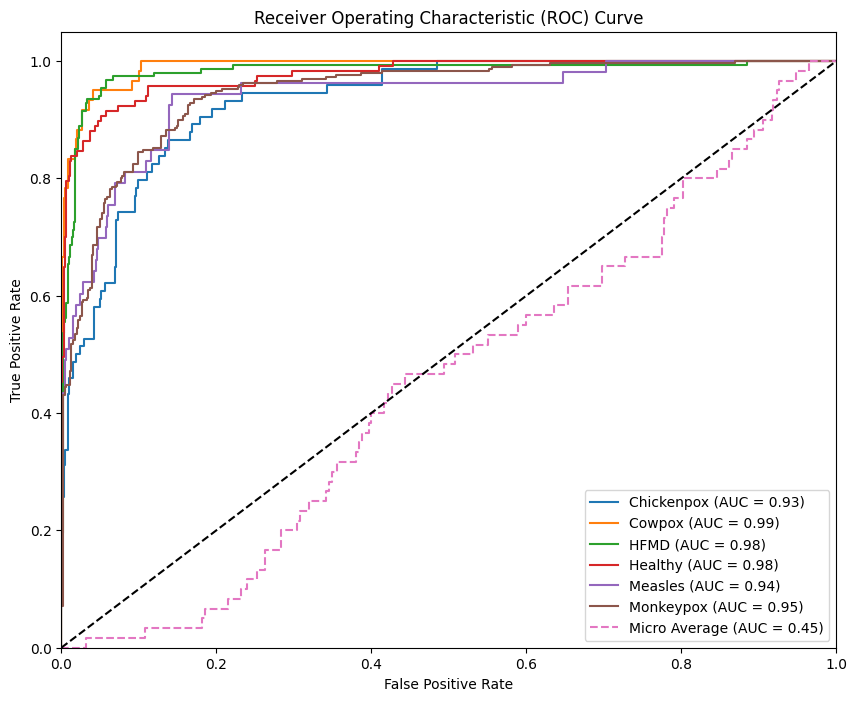}
            \caption{VGG19}
            \label{vgg19_mul_roc}
            
        \end{subfigure}

        \vspace{1em}

        \begin{subfigure}[b]{0.45\textwidth}
            \includegraphics[width=\linewidth]{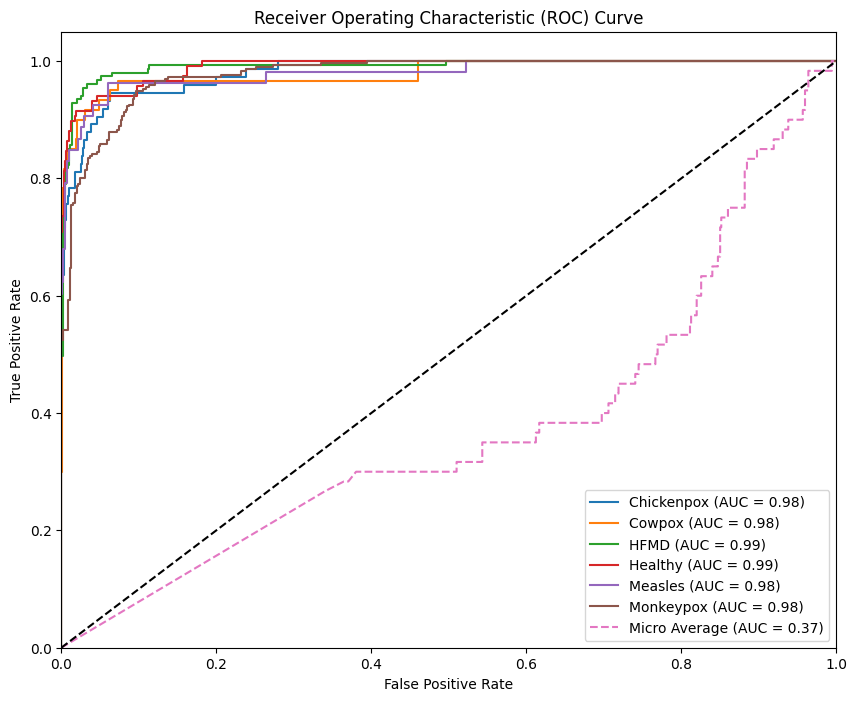}
            \caption{Inception}
            \label{incep_mul_roc}
        \end{subfigure}
        \hspace{1em}
        \begin{subfigure}[b]{0.45\textwidth}
            \includegraphics[width=\linewidth]{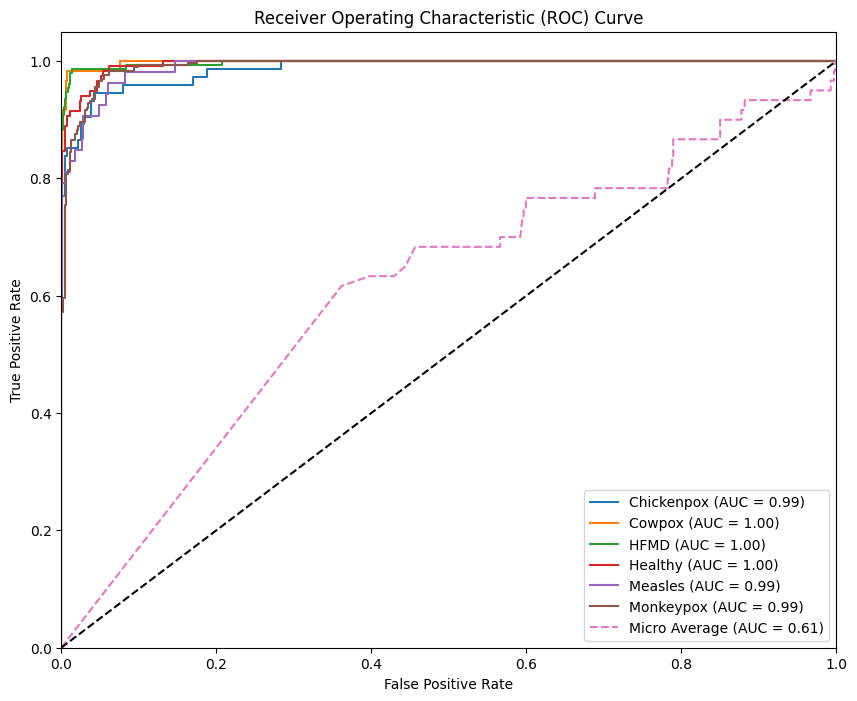}
            \caption{MobileNetV2}
            \label{mobile_mul_roc}
        \end{subfigure}

    \vspace{2em}

   
    \caption{Visual comparison of model performance: ROC curves }
    \label{fig:roc_mul}
\end{figure*}

\begin{figure*}[t]
    \centering

    \begin{subfigure}[b]{0.45\textwidth}
        \centering
        \includegraphics[width=\linewidth]{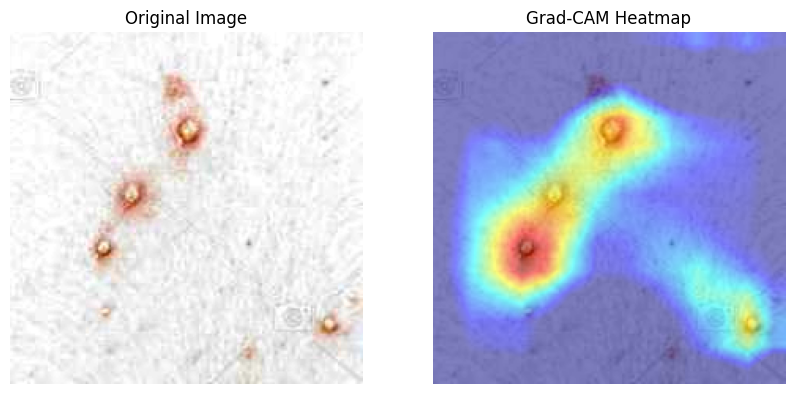}
        \caption{VGG16 Grad-CAM}
        \label{fig:gradcam_vgg16}
    \end{subfigure}
    \hfill
    \begin{subfigure}[b]{0.45\textwidth}
        \centering
        \includegraphics[width=\linewidth]{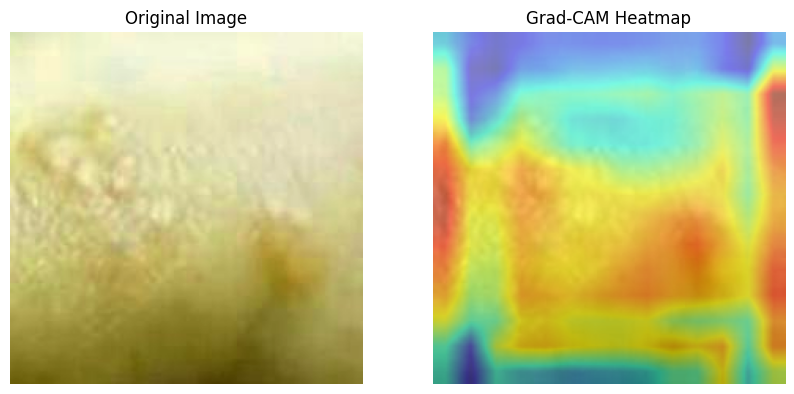}
        \caption{VGG19 Grad-CAM}
        \label{fig:gradcam_vgg19}
    \end{subfigure}

    \vspace{0.5em}

    \begin{subfigure}[b]{0.45\textwidth}
        \centering
        \includegraphics[width=\linewidth]{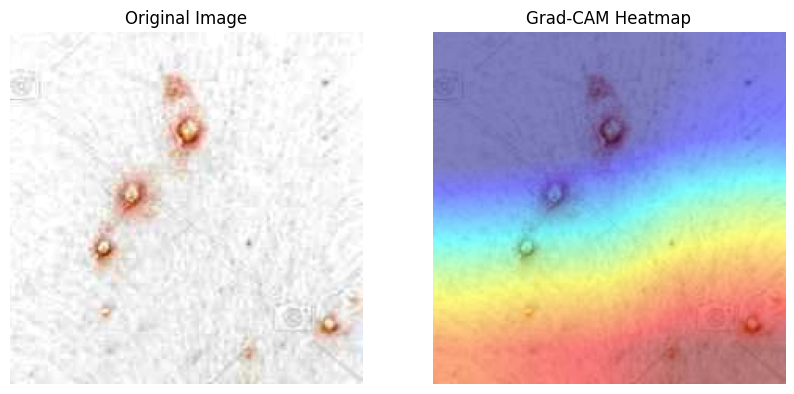}
        \caption{Inception Grad-CAM}
        \label{fig:gradcam_incep}
    \end{subfigure}
    \hfill
    \begin{subfigure}[b]{0.45\textwidth}
        \centering
        \includegraphics[width=\linewidth]{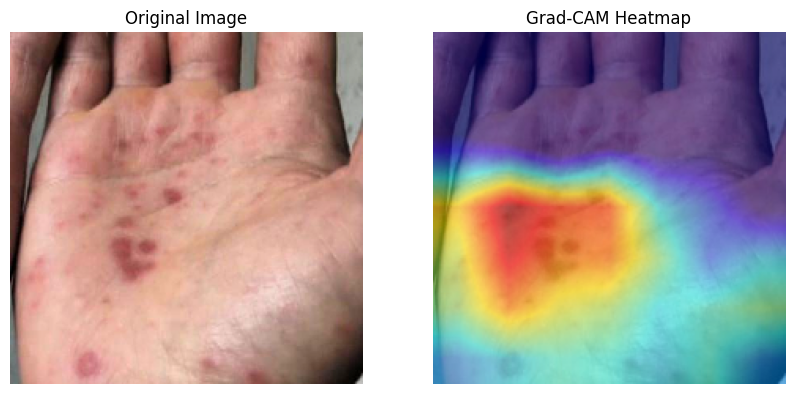}
        \caption{MobileNetV2 Grad-CAM}
        \label{fig:gradcam_mobilenet}
    \end{subfigure}

    \caption{Grad-CAM visualizations for the four models for multi class.}
    \label{fig:gradcam_all_mul}
\end{figure*}

Inception-V3 achieves superior results compared to both VGG models, with accuracy, precision, recall, and F1-score of 0.89, 0.91, 0.85, and 0.88, respectively. As shown in \textbf{Figure \ref{fig:confusion_mul}}, it makes only 87 incorrect predictions while correctly classifying 667 cases. The loss score of 0.55 is higher than VGG-16 but is mitigated by better prediction metrics. Despite its better performance, the wall time matches VGG-16 at 1 hour 44 minutes. The CPU time is also the same at 14 minutes 54 seconds, highlighting its efficient architecture. Inception-V3 outperforms VGG-16 and VGG-19 with higher precision and recall while maintaining similar computational times.
MobileNetV2 delivers the best performance among all models, with accuracy, precision, recall, and F1-score of 0.93, 0.93, 0.90, and 0.91, respectively. Its loss score of 0.32 is the lowest, indicating exceptional generalization. As shown in \textbf{Figure \ref{fig:cm4}}, it makes only 52 incorrect predictions, which is fewer than the VGG and Inception-V3. The wall time is significantly lower at 56 minutes 58 seconds, making it the fastest model. Similarly, its CPU time is the lowest at 6 minutes 9 seconds, reflecting its lightweight and efficient design. MobileNetV2 outperforms all other models in both accuracy and computational efficiency.

\textbf{Analysis of Accuracy and loss curves}:
\textbf{Figure \ref{fig:all_acc_loss_mul}} displays accuracy and loss of all four models for multiclass dataset.
The training and validation curves for VGG16; \textbf{Figure \ref{vgg16_mul_acc}} demonstrate strong and consistent performance. Training accuracy steadily increases, reaching approximately 0.95, while validation accuracy stabilizes around 0.87 by the final epochs, indicating robust learning and generalization. The training loss decreases consistently to about 0.2, and the validation loss plateaus at approximately 0.4, reflecting good convergence between the two metrics. These results suggest that VGG16 effectively learns the data patterns with minimal overfitting, making it the most generalized model in this analysis.
In \textbf{Figure \ref{vgg19_mul_acc}} training and validation curves for VGG19 are shown. The curve shows a similar trend to VGG16 but with slightly lower validation performance. Training accuracy gradually improves, peaking at around 0.93, while validation accuracy stabilizes at approximately 0.85, indicating effective learning but slightly weaker generalization compared to VGG16. The training loss steadily declines to around 0.3, and the validation loss stabilizes at approximately 0.4, showing a modest gap between the metrics. These trends suggest that while VGG19 learns effectively, it may slightly overfit compared to VGG16, as seen in the slightly lower validation accuracy.
For InceptionV3 \textbf{Figure \ref{incep_mul_acc}} , the model was trained for 30 epochs, with the last 10 epochs displayed after resuming from a saved checkpoint. Training accuracy continues to rise in these final epochs, reaching 0.93, while validation accuracy fluctuates around 0.91, indicating some instability in generalization. Training loss steadily decreases to approximately 0.15, while validation loss remains higher, ranging between 0.45 and 0.5, suggesting mild overfitting. These results indicate that InceptionV3 optimizes well on training data but struggles with consistent validation performance, requiring additional regularization or fine-tuning to enhance generalization.
\textbf{Figure \ref{mobile_mul_acc}}. illustrates the training and validation accuracy and loss for MobileNetV2 during the last 10 epochs of training. The initial training consisted of 20 epochs, followed by saving the model and continuing with early stopping applied. The training accuracy shows a steady increase, nearing 1.0, while the validation accuracy stabilizes above 0.92 with minor fluctuations, indicating good generalization. Training loss decreases consistently, reaching minimal values, while validation loss fluctuates slightly but trends downward, aligning with the accuracy improvements. Overall, the model demonstrates effective learning and stable performance on unseen data.\\

\textbf{Analysis of Roc curves}:
Figure \ref{vgg16_mul_roc} and Figure \ref{vgg19_mul_roc} illustrate the ROC curves for the VGG16 and VGG19 models, respectively, evaluating classification performance across six classes: Chickenpox, Cowpox, HFMD, Healthy, Measles, and Monkeypox. Both models demonstrate strong discriminative capabilities, with VGG16 achieving the highest AUC of 0.99 for Cowpox, HFMD, and Healthy, followed by Chickenpox and Monkeypox at 0.97, and Measles at 0.95. Similarly, VGG19 also performs well, with Cowpox achieving an AUC of 0.99, HFMD and Healthy at 0.98, Monkeypox at 0.95, Measles at 0.94, and Chickenpox at 0.93. However, the micro-average AUC is lower for both models, at 0.47 for VGG16 and 0.45 for VGG19, indicating reduced aggregated performance across all classes. Overall, both models exhibit high classification accuracy for individual classes, with VGG16 slightly outperforming VGG19 in some categories.
In Figure \ref{incep_mul_roc}, the ROC curve for the InceptionV3 model demonstrates high per-class AUC values (~0.98–0.99), indicating strong discrimination for individual classes such as Chickenpox, Cowpox, and Measles. However, the micro-average AUC is notably low at 0.37, suggesting the model struggles to maintain consistent performance across all classes when aggregating predictions, likely due to class imbalance or errors in some predictions. 
In contrast, Figure \ref{mobile_mul_roc} shows the ROC curve for the MobileNetV2 model, which achieves higher per-class AUCs (up to 1.00 for some classes like Cowpox and HFMD) and a significantly better micro-average AUC of 0.61. This indicates that MobileNetV2 provides more reliable and consistent performance across all classes, outperforming InceptionV3 in terms of overall classification ability.

\textbf{Feature highlights using Grad-CAM for multiclass data}
In Figure \ref{fig:gradcam_all_mul}, some samples of the predicted heatmap generated by GRAD-CAM are shown of the four pretrained models that we used for multi-class dataset.\\

\textbf{RQ1 Summary}: For binary class data, InceptionV3 demonstrates the best performance with 0.95 accuracy, precision, recall, and F1-score. However, MobileNetV2 outperforms all models in terms of wall-clock time and CPU efficiency. Conversely, for multi-class data, MobileNetV2 achieves the highest performance across all metrics, with 0.93 accuracy and precision, and a CPU time of 6 minutes and 9 seconds.\\
\textbf{RQ4 Summary}: To interpret the performance of the CNN model, an Explainable AI (XAI) technique called GRAD-CAM is applied. GRAD-CAM is a visualization method that highlights the important features of a test image using a heatmap. This technique enhances the model's reliability and trustworthiness.\\
\textbf{Summary} 
The results show that InceptionV3 performed best on the binary dataset with 95\% accuracy, while MobileNetV2 achieved the highest accuracy (93\%) on the multi-class dataset, demonstrating its computational efficiency. Grad-CAM provided effective visualizations of key features contributing to predictions, enhancing interpretability. Despite high performance, some models exhibited overfitting, as seen in discrepancies between training and validation losses. Overall, the study highlights the effectiveness of pre-trained CNN models and the value of XAI techniques in improving diagnostic accuracy and trustworthiness.

\section{Discussion}
This study explored the efficacy of pre-trained Convolutional Neural Network (CNN) models, including VGG16, VGG19, InceptionV3, and MobileNetV2, for early monkeypox detection using two distinct datasets: a binary-class dataset and a multi-class dataset (confusion matrices in figure 7 and figure 10). Each dataset was augmented to overcome the challenges of limited medical imaging data and to enhance model generalization.\\
\textbf{Dataset Performance and Overfitting Concerns}
The results indicate significant differences in model performance between the binary and multi-class datasets. For the binary-class dataset, InceptionV3 and MobileNetV2 outperformed VGG16 and VGG19, achieving accuracy rates of 95\% and 94\%, respectively. On the multi-class dataset, MobileNetV2 exhibited superior performance with an accuracy of 93\%, followed by InceptionV3 at 89\%. The VGG models lagged slightly, likely due to their more complex architectures that are less suited for smaller datasets.
Overfitting emerged as a key concern, particularly with InceptionV3 and VGG19. While training losses approached near-zero values, validation losses plateaued at higher levels, suggesting the models may have learned patterns specific to the training data rather than generalizing effectively. The use of early stopping and dropout layers mitigated this risk to some extent, as evidenced by stabilized validation accuracy curves. Data augmentation also played a pivotal role in addressing overfitting by increasing the diversity of training samples.\\

\textbf{Comparative Analysis of Metrics}
The models were evaluated based on several metrics, including accuracy, precision, recall, F1-score, loss, wall time, and CPU time. MobileNetV2 consistently demonstrated the best balance between performance and computational efficiency. For instance, its training wall time was significantly lower than other models, at 6 minutes 48 seconds for the binary-class dataset and 56 minutes 58 seconds for the multi-class dataset. In contrast, the VGG models required over an hour for multi-class training, highlighting their computational intensity.

\begin{table}[htbp]
    \centering
    \caption{Comparison with other state-of-the-art works (Part 1: Core Metrics)}
    \label{tab:comparison_sota_part1}
    \renewcommand{\arraystretch}{1.2}

    \begin{tabular}{|p{1.8cm}|p{1.8cm}|p{2.4cm}|c|c|c|}
        \hline
        \textbf{Study} & \textbf{Dataset validated} & \textbf{Best model name} 
        & \textbf{Accuracy.} & \textbf{Precision.} & \textbf{Recall} \\
        \hline
        \citep{bib14} & One-binary & Modified-VGG19 & 0.93 & 0.94 & 0.94 \\
        \hline
        \citep{bib24} & One-binary & Ensemble & 0.87 & 0.85 & 0.85 \\
        \hline
        \citep{bib44} & One-multi & Vision-CNN & 0.81 & 0.83 & 0.85 \\
        \hline
        \multirow{2}{*}{Our Study} & Both & InceptionV3-binary & 0.95 & 0.95 & 0.95 \\
        \cline{2-6}
         &  & MobileNetV2-Multi & 0.93 & 0.93 & 0.90 \\
        \hline
    \end{tabular}
\end{table}

\begin{table}[htbp]
    \centering
    \caption{Comparison with other state-of-the-art works (Part 2: Other Metrics and Explainability)}
    \label{tab:comparison_sota_part2}
    \renewcommand{\arraystretch}{1.2}

    \begin{tabular}{|p{1.8cm}|p{1.8cm}|p{2.4cm}|c|c|p{1.5cm}|p{1.5cm}|p{1cm}|}
        \hline
        \textbf{Study} & \textbf{Dataset validated} & \textbf{Best model name} 
        & \textbf{Loss} & \textbf{F1} & \textbf{Wall time} & \textbf{CPU time} & \textbf{XAI} \\
        \hline
        \citep{bib14} & One-binary & Modified-VGG19 & N/A & 0.94 & N/A & N/A & LIME \\
        \hline
        \citep{bib24} & One-binary & Ensemble & N/A & 0.85 & N/A & N/A & Grad-CAM, LIME \\
        \hline
        \citep{bib44} & One-multi & Vision-CNN & N/A & 0.86 & N/A & N/A & N/A \\
        \hline
        \multirow{2}{*}{Our Study} & Both & InceptionV3-binary & 0.11 & 0.95 & 15m 50s & 3m 16s & Grad-CAM \\
        \cline{2-8}
         &  & MobileNetV2-Multi & 0.32 & 0.91 & 56m 58s & 6m 9s & \\
        \hline
    \end{tabular}
\end{table}

The ROC and confusion matrix analyses further confirmed the reliability of MobileNetV2. It achieved the highest Area Under the Curve (AUC) values across most classes and made fewer incorrect predictions compared to other models. This suggests that its lightweight architecture is particularly well-suited for applications requiring rapid and accurate 
predictions.\\
\textbf{Explainable AI (XAI) Insights}
The integration of Gradient-weighted Class Activation Mapping (Grad-CAM) provided critical insights into model interpretability. Grad-CAM heatmaps highlighted the key regions in images that influenced predictions, such as skin lesions specific to monkeypox or other classes. This not only enhanced the transparency of model decisions but also validated the clinical relevance of predictions, making the approach more trustworthy for healthcare applications.\\
\textbf{Comparison with other State of the Art}
\textbf{Table \ref{tab:comparison_sota_part1}} and  \textbf{Table \ref{tab:comparison_sota_part2}} highlight the comparative performance of various studies and models, emphasizing the strengths of our work against other authors' contributions. Our InceptionV3-binary achieved the highest accuracy and F1-score of 0.95, outperforming Ahsan’s Modified-VGG19, which recorded an accuracy of 0.93, and Chi’s Ensemble model, which achieved only 0.87. In multi-class classification, our Modified MobileNetV2 achieved an accuracy of 0.93, which is notably higher than the 0.81 accuracy reported for Vision-CNN by \citep{bib44}.Additionally, we uniquely reported Wall time (e.g., 15m 50s for InceptionV3-binary and 56m 58s for MobileNetV2) and CPU time, metrics not addressed by \citep{bib14}, \citep{bib24}, or \citep{bib44}. Moreover, we utilized Grad-CAM for explainable AI, enhancing model interpretability, while Ahsan used LIME and Chi combined Grad-CAM and LIME. Unlike these authors, who primarily evaluated their models on single datasets, our use of diverse datasets adds robustness and sets a higher standard for future research.\\
\textbf{Deployment}
We propose deploying the best-performing deep learning model as a user-friendly tool with built-in Grad-CAM visualizations for interpretability. The model can be containerized and hosted on a secure server or integrated into a web app to provide real-time predictions. This setup would help end-users apply the system in practice while understanding its decisions.

\section{Threats to Validity}
\textbf{Internal validity}
Internal validity refers to the extent to which the experimental design ensures that the observed effects are due to the independent variables rather than extraneous factors. In this study, pre-trained models (e.g., VGG16, VGG19, Inception V3, Mobile Net V2) were fine-tuned on datasets specifically to the monkey pox dataset. However, these models were originally trained on ImageNet (dataset that contains non-medical images). This may lead to biases as the features learned during pretraining may not fully align with the nuances of medical imaging. To minimize the effects of this, customized hyperparameters tunning was done and good results were found.
Moreover, the data augmentation might introduce artificial patterns which may not be in the real-world scenarios. Such a factor could impact the model’s ability to generalize across real-world applications \citep{bib45}. \\
\textbf{External validity}
External validity addresses the generalizability of the findings to other contexts or populations \citep{bib46}. In this study, MSLD and MSLD V2 were used for the experiment, that may not capture the full variability of disease in the diversity of the population. Moreover, dataset’s image conditions might not reflect real-world environment because of the lighting, camera type and resolution. This is enough to limit the applicability of the result beyond the score of the current datasets. However, in this study we tried to overcome this condition by using augmented images for ensuring diversity of the categories. \\
\textbf{Construct validity}
In this study, the performance metrics (accuracy, precision, recall, and F1 score) provide a standard view of model’s effectiveness. However, for datasets with class imbalances, metrics such as Area Under the Precision-Recall Curve (PR-AUC) or Matthews Correlation Coefficient (MCC) could provide more robust insights. Relying only on traditional metrics may overlook the model’s performance especially for underrepresented classes \citep{bib47}.\\
\textbf{Conclusion validity}
Relatively small dataset size and imbalance in class distribution may introduce statistical biases, which reduce the reliability of the reported matric \citep{bib48}. Additionally, while hyperparameters such as batch size and learning rate are carefully tuned, the outcomes might vary under different datasets. However, this study utilizes augmented images where enough images were used to reduce statistical biases. 
\section{Future work}
The first limitation of our work is the limited dataset used for validation, as we relied on only two publicly available datasets. Additionally, our model produced some false positive and false negative predictions, which is a critical concern in the medical field where reducing false predictions is essential. We employed Grad-CAM to interpret the model’s performance. However, the relationship between Grad-CAM outputs and the CNN model’s performance was not explicitly established, leaving the relevance of CNN predictions and Grad-CAM visualizations unclear.
There is significant room for improvement, providing opportunities for future researchers to address these issues. One potential direction for future work is the use of multimodal datasets—combining image data with patients’ clinical data—to achieve more accurate predictions. Furthermore, our study utilized only one interpretability technique (Grad-CAM). Exploring other techniques, such as Grad-CAM++, LIME, or SHAP, could enhance the transparency of the model. Another promising avenue is to establish a clear mapping between CNN performance and model interpretability, which would help healthcare practitioners make more informed decisions regarding patient care.
\section{Conclusion}
The main objective of our research was the early detection of mpox using skin-lesion images using pre-trained CNN models and leveraging XAI technique to interpret the model results. Three research questions are developed and answered throughout the whole paper. Datasets: Two publicly available datasets were used for monkeypox detection: one for binary classification and another for multi-class classification. This approach validates the model's robustness and ensures generalizability across different classification scenarios.
\textbf{Proposed model}: Transfer learning techniques were incorporated using four pre-trained models: VGG16, VGG19, InceptionV3, and MobileNetV2. These models were fine-tuned by freezing the initial layers and adding additional layers to enhance feature extraction and make the models more lightweight.
\textbf{Hyperparameters}: Adam was chosen as the optimizer, categorical cross entropy as the loss function, the learning rate was set to 0.0001, and early stopping was implemented to avoid unnecessary computations. Although the total number of epochs was set to 30, all models stopped training early due to the early stopping mechanism, which halted training after performance failed to improve for 5 consecutive epochs.
\textbf{Results and analysis}: InceptionV3 gave the highest performance in terms of accuracy, precision, recall and F1-score of 0.95 in the binary dataset. However, mobilenetV2’s wall time and CPU time was better. On the other hand, for multi-class data MobileNetV2 performed the best showing an accuracy of 0.93. However, by observing accuracy and loss curve it is clearly visible that there are some overfitting issues in most of the models.
\textbf{Explainable AI}: To interpret the model's performance, Grad-CAM was used to highlight the key features contributing to the prediction of a specific class. This interpretability is crucial as it helps practitioners build trust in the AI system.
This study provides valuable insights into the performance of pre-trained CNN models for the early detection of the monkeypox virus, while also enhancing model transparency using XAI techniques. Future researchers can develop more robust models to assist practitioners in making informed decisions about the disease.

\section{Dataset Availability:}
The datasets are publicly available in kaggle:
\textbf{ln.run/oHLqJ}


\begin{thebibliography}{00}

\bibitem{bib1}
Le´on-Figueroa, D.A., Barboza, J.J., Garcia-Vasquez, E.A., Bonilla-Aldana,
D.K., Diaz-Torres, M., Salda˜na-Cumpa, H.M., Diaz-Murillo, M.T., Cruz,
O.C.S., Rodriguez-Morales, A.J., 2022. \textit{Epidemiological situation of mon-
keypox transmission by possible sexual contact: a systematic review.} Trop-
ical Medicine and Infectious Disease 7, 267.https://doi.org/10.3390/tropicalmed7100267 



\bibitem{bib2}

de Noordhout, C.M., Devleesschauwer, B., Haagsma, J.A., Havelaar, A.H.,
Bertrand, S., Vandenberg, O., Quoilin, S., Brandt, P.T., Speybroeck, N.,
2017. \textit{Burden of salmonellosis, campylobacteriosis and listeriosis: a time
series analysis, belgium, 2012 to 2020}. Eurosurveillance 22, 30615.
https://doi.org/10.2807/1560-7917 

\bibitem{bib3}
Kugelman, J.R., Johnston, S.C., Mulembakani, P.M., Kisalu, N., Lee, M.S.,
Koroleva, G., McCarthy, S.E., Gestole, M.C., Wolfe, N.D., Fair, J.N., et al.,
2014. \textit{Genomic variability of monkeypox virus among humans, democratic
republic of the congo.} Emerging infectious diseases 20, 232.



\bibitem{bib4}

Kaler, J., Hussain, A., Flores, G., Kheiri, S., Desrosiers, D., 2022.\textit{Monkeypox:
a comprehensive review of transmission, pathogenesis, and manifestation}.
Cureus 14.

\bibitem{bib5}
Xiang, Y., White, A., 2022. \textit{Monkeypox virus emerges from the shadow of
its more infamous cousin: family biology matters}. Emerging microbes \&
infections 11, 1768–1777.https://doi.org/10.1080/22221751.2022.2095309 



\bibitem{bib6}

Saxena, S.K., Ansari, S., Maurya, V.K., Kumar, S., Jain, A., Paweska, J.T., Tri-
pathi, A.K., Abdel-Moneim, A.S., 2023. \textit{Re-emerging human monkeypox:
a major public-health debacle.} Journal of medical virology 95, e27902. https://doi.org/10.1002/jmv.27902 

\bibitem{bib7}

 Farahat, R.A., Sah, R., El-Sakka, A.A., Benmelouka, A.Y., Kundu, M., Labieb,
F., Shaheen, R.S., Abdelaal, A., Abdelazeem, B., Bonilla-Aldana, D.K.,
et al., 2022.\textit{ Human monkeypox disease (mpx)}. Le infezioni in Medicina
30, 372.https://doi.org/10.53854/liim-3003-6 

\bibitem{bib8}

Sam-Agudu, N.A., Martyn-Dickens, C., Ewa, A.U., 2023. \textit{A global update of
mpox (monkeypox) in children}. Current Opinion in Pediatrics 35, 193–200.

\bibitem{bib9}
Yinka-Ogunleye, A., Aruna, O., Dalhat, M., Ogoina, D., McCollum, A., Disu,
Y., Mamadu, I., Akinpelu, A., Ahmad, A., Burga, J., et al., 2019. \textit{Outbreak
of human monkeypox in nigeria in 2017–18: a clinical and epidemiological
report.} The Lancet Infectious Diseases 19, 872–879.https://doi.org/10.1016/S1473-3099(19)30294-4 



\bibitem{bib10}
Kumar, A., Borkar, S.K., Choudhari, S.G., Mendhe, H.G., Bankar, N.J.,
Mendhe, H., 2023. \textit{ Recent outbreak of monkeypox: implications for public
health recommendations and crisis management in india.} Cureus 15.

\bibitem{bib11}
Thornhill, J.P., Barkati, S., Walmsley, S., Rockstroh, J., Antinori, A., Harrison,
L.B., Palich, R., Nori, A., Reeves, I., Habibi, M.S., et al., 2022. \textit{Monkey-
pox virus infection in humans across 16 countries—april–june 2022.} New
England Journal of Medicine 387, 679–691. https://doi.org/10.1056/nejmoa2207323 

\bibitem{bib12}

Gong, Q., Wang, C., Chuai, X., Chiu, S., 2022. \textit{Monkeypox virus: a re-
emergent threat to humans}. Virologica Sinica 37, 477–482.https://doi.org/10.1016/j.virs.2022.07.006
 





\bibitem{bib13}
Agrebi, S., Larbi, A., 2020. \textit{Use of artificial intelligence in infectious diseases},
in: Artificial intelligence in precision health. Elsevier, pp. 415–438.https://doi.org/10.1016/B978-0-12-817133-2.00018-5 


\bibitem{bib14}
 Ahsan, M.M., Luna, S.A., Siddique, Z., 2022. \textit{Machine-learning-based disease
diagnosis: A comprehensive review}, in: Healthcare, MDPI. p. 541.https://doi.org/10.3390/healthcare10030541

 \bibitem{bib15}
Van der Velden, B.H., Kuijf, H.J., Gilhuijs, K.G., Viergever, M.A., 2022.\textit{ Ex-
plainable artificial intelligence (xai) in deep learning-based medical image
analysis}. Medical Image Analysis 79, 102470.https://doi.org/10.1016/j.media.2022.102470 


\bibitem{bib16}
Zuluaga-Gomez, J., Al Masry, Z., Benaggoune, K., Meraghni, S., Zerhouni, N.,
2021. \textit{A cnn-based methodology for breast cancer diagnosis using thermal
images.} Computer Methods in Biomechanics and Biomedical Engineering:
Imaging \& Visualization 9, 131–145.https://doi.org/10.1080/21681163.2020.1824685

\bibitem{bib17}
Li, M., Jiang, Y., Zhang, Y., Zhu, H., 2023. \textit{Medical image analysis using deep
learning algorithms.} Frontiers in public health 11, 1273253.
https://doi.org/10.3389/fpubh.2023.1273253

\bibitem{bib18}

Kim, H.E., Cosa-Linan, A., Santhanam, N., Jannesari, M., Maros, M.E., Gans-
landt, T., 2022. \textit{Transfer learning for medical image classification: a litera-
ture review.} BMC medical imaging 22, 69.https://doi.org/10.1186/s12880-022-00793-7



\bibitem{bib19}


He, K., Gan, C., Li, Z., Rekik, I., Yin, Z., Ji, W., Gao, Y., Wang, Q., Zhang,
J., Shen, D., 2023.\textit{Transformers in medical image analysis.} Intelligent
Medicine 3, 59–78.https://doi.org/10.1016/j.imed.2022.07.002 

\bibitem{bib20}

Wang, J., Perez, L., et al., 2017.\textit{The effectiveness of data augmentation in
image classification using deep learning.} Convolutional Neural Networks
Vis. Recognit 11, 1–8.


\bibitem{bib21}
Yasmin, F., Hassan, M.M., Hasan, M., Zaman, S., Kaushal, C., El-Shafai, W.,
Soliman, N.F., 2023.\textit{Poxnet22: A fine-tuned model for the classification of
monkeypox disease using transfer learning.} Ieee Access 11, 24053–24076.https://doi.org/10.1109/ACCESS.2023.3253868 




 \bibitem{bib22}
 Ahsan, M.M., Ali, M.S., Hassan, M.M., Abdullah, T.A., Gupta, K.D., Bagci,
U., Kaushal, C., Soliman, N.F., 2023. \textit{Monkeypox diagnosis with inter-
pretable deep learning.} IEEE Access 11, 81965–81980.https://doi.org/10.1016/j.eswa.2022.119483 


\bibitem{bib23}

Raha, A.D., Gain, M., Debnath, R., Adhikary, A., Qiao, Y., Hassan, M.M.,
Bairagi, A.K., Islam, S.M.S., 2024. \textit{Attention to monkeypox: An inter-
pretable monkeypox detection technique using attention mechanism.} IEEE
Access .https://doi.org/10.1109/ACCESS.2024.3385099 


\bibitem{bib24}

Sitaula, C., Shahi, T.B., 2022. \textit{Monkeypox virus detection using pre-trained
deep learning-based approaches.} Journal of Medical Systems 46, 78.https://doi.org/10.1007/s10916-022-01868-2 

\bibitem{bib25}
Sorayaie Azar, A., Naemi, A., Babaei Rikan, S., Bagherzadeh Mohasefi, J.,
Pirnejad, H., Wiil, U.K., 2023.\textit{ Monkeypox detection using deep neural
networks.} BMC Infectious Diseases 23, 438.https://doi.org/10.1186/s12879-023-08408-4 


\bibitem{bib26}
Dahiya, N., Sharma, Y.K., Rani, U., Hussain, S., Nabilal, K.V., Mohan, A.,
Nuristani, N., 2023.\textit{ Hyper-parameter tuned deep learning approach for ef-
fective human monkeypox disease detection}. Scientific reports 13, 15930.https://doi.org/10.1038/s41598-023-43236-1 


\bibitem{bib27}
Yolcu Oztel, G., 2024.\textit{ Vision transformer and cnn-based skin lesion analysis:
classification of monkeypox.} Multimedia Tools and Applications 83, 71909–
71923.https://doi.org/10.1007/s11042-024-19757-w 


\bibitem{bib28}

Singh, C., Ranade, S.K., Singh, S.P., 2024.\textit{Attention learning models using
local zernike moments-based normalized images and convolutional neural
networks for skin lesion classification.} Biomedical Signal Processing and
Control 96, 106512.https://doi.org/10.1016/j.bspc.2024.106512 


\bibitem{bib29}
Maqsood, S., Damaˇseviˇcius, R., Shahid, S., Forkert, N.D., 2024.\textit{ Mox-net:
Multi-stage deep hybrid feature fusion and selection framework for mon-
keypox classification.} Expert Systems with Applications 255, 124584. https://doi.org/10.1016/J.ESWA.2024.124584

\bibitem{bib30}
Bala, D., Hossain, M. S., Hossain, M. A., Abdullah, M. I., Rahman, M. M., Manavalan, B., Gu, N., Islam, M. S., \& Huang, Z. (2023). \textit{MonkeyNet: A robust deep convolutional neural network for monkeypox disease detection and classification.} Neural Networks, 161, 757–775.https://doi.org/10.1016/j.neunet.2023.02.022 

\bibitem{bib31}
Asif, S., Zhao, M., Tang, F., Zhu, Y., Zhao, B., 2023. \textit{Metaheuristics
optimization-based ensemble of deep neural networks for mpox disease de-
tection.} Neural Networks 167, 342–359.https://doi.org/10.1016/j.neunet.2023.08.035 


\bibitem{bib32} Oztel, I., Oztel, G. Y.,\& Sahin, V. H. (2023).\textit{Deep Learning-Based Skin Diseases Classification using Smartphones.} Advanced Intelligent Systems, 5(12). https://doi.org/10.1002/aisy.202300211 


\bibitem{bib33}
Caldiera, V.R.
B.G., Rombach, H.D., 1994. \textit{The goal question metric approach.}
Encyclopedia of software engineering , 528–532.






\bibitem{bib34}

Wang, J., Perez, L., et al., 2017. \textit{The effectiveness of data augmentation in
image classification using deep learning.} Convolutional Neural Networks
Vis. Recognit 11, 1–8.

\bibitem{bib35}
Taylor, L., Nitschke, G., 2018.\textit{ Improving deep learning with generic data aug-
mentation}, in: 2018 IEEE symposium series on computational intelligence
(SSCI), IEEE. pp. 1542–1545.https://doi.org/10.1109/SSCI.2018.8628742 


\bibitem{bib36}
  Ali, S.N., Ahmed, M.T., Jahan, T., Paul, J., Sani, S.S., Noor, N., Asma, A.N.,
Hasan, T., 2024. \textit{A web-based mpox skin lesion detection system using
state-of-the-art deep learning models considering racial diversity}. Biomedi-
cal Signal Processing and Control 98, 106742.https://arxiv.org/abs/2306.14169v1

\bibitem{bib37}

Simonyan, K., Zisserman, A., 2014.\textit{Very deep convolutional networks for
large-scale image recognition.} arXiv preprint arXiv:1409.1556 .

\bibitem{bib38}

Sandler, M., Howard, A., Zhu, M., Zhmoginov, A., Chen, L.C., 2018.\textit{Mo-
bilenetv2: Inverted residuals and linear bottlenecks}, in: Proceedings of the
IEEE conference on computer vision and pattern recognition, pp. 4510–
4520.https://doi.org/10.1109/CVPR.2018.00474 

\bibitem{bib39}
Szegedy, C., Vanhoucke, V., Ioffe, S., Shlens, J., Wojna, Z., 2016.\textit{Rethinking
the inception architecture for computer vision}, in: Proceedings of the IEEE conference on computer vision and pattern recognition, pp. 2818–2826.https://doi.org/10.1109/CVPR.2016.308 




 \bibitem{bib40}
 Ahsan, M.M., Li, Y., Zhang, J., Ahad, M.T., Yazdan, M.M.S., 2020. \textit{Face
recognition in an unconstrained and real-time environment using novel bmc-
lbph methods incorporates with dji vision sensor}. Journal of Sensor and
Actuator Networks 9, 54.https://doi.org/10.3390/jsan9040054 

\bibitem{bib41}
Bottou, L., Curtis, F.E., Nocedal, J., 2018.\textit{ Optimization methods for large-
scale machine learning}. SIAM review 60, 223–311.

\bibitem{bib42}

Kingma, D.P., Ba, J., 2015. \textit{Adam: A method for stochastic optimization. in-
ternational conference on learning representations (2015)}. San Diego, Cali-
fornia .https://arxiv.org/abs/1412.6980v9 
\bibitem{bib43}
Selvaraju, R. R., Cogswell, M., Das, A., Vedantam, R., Parikh, D., \& Batra, D. (2016).\textit{Grad-CAM: Visual Explanations from Deep Networks via Gradient-based Localization.} https://doi.org/10.1007/s11263-019-01228-7 

\bibitem{bib44}

Yolcu Oztel, G., 2024.\textit{Vision transformer and cnn-based skin lesion analysis:
classification of monkeypox.} Multimedia Tools and Applications 83, 71909–
71923. https://doi.org/10.1007/s11042-024-19757-w 

\bibitem{bib45}

Perez, L., Wang, J., 2017.\textit{The effectiveness of data augmentation in image
classification using deep learning.} arXiv preprint arXiv:1712.04621 .

\bibitem{bib46}

Findley, M.G., Kikuta, K., Denly, M., 2021. \textit{External validity.} Annual review
of political science 24, 365–393.https://doi.org/10.1146/annurev-polisci-041719-102556 

\bibitem{bib47}

Saito, T., Rehmsmeier, M., 2015. \textit{The precision-recall plot is more informative
than the roc plot when evaluating binary classifiers on imbalanced datasets.}
PloS one 10, e0118432. https://doi.org/10.1371/JOURNAL.PONE.0118432 

\bibitem{bib48}

Ghosh, K., Bellinger, C., Corizzo, R., Branco, P., Krawczyk, B., Japkowicz,
N., 2024. \textit{The class imbalance problem in deep learning}. Machine Learning
113, 4845–4901. https://doi.org/10.1007/S10994-022-06268-8/FIGURES/27 

\bibitem{bib49}
Ali, S. N., Ahmed, Md. T., Jahan, T., Paul, J., Sani, S. M. S., Noor, N., Asma, A. N., \& Hasan, T. (2023). \textit{A Web-based Mpox Skin Lesion Detection System Using State-of-the-art Deep Learning Models Considering Racial Diversity}. https://arxiv.org/abs/2306.14169v1 

\bibitem{bib50}
Ali, S. N., Ahmed, Md. T., Paul, J., Jahan, T., Sani, S. M. S., Noor, N., \& Hasan, T. (2022).\textit{Monkeypox Skin Lesion Detection Using Deep Learning Models: A Feasibility Study}. https://arxiv.org/abs/2207.03342v1



 

 \

















 
 


\end{thebibliography}
\end{document}